%% file: main.tex
\documentclass[10pt,twocolumn,letterpaper]{article}

\usepackage{cvpr}
\usepackage{times}
\usepackage{epsfig}
\usepackage{graphicx}
\usepackage{amsmath}
\usepackage{amssymb}

\usepackage{subcaption}
\usepackage{booktabs}
\usepackage{multirow}
\usepackage{bm}
\usepackage[inline]{enumitem}
\usepackage{amsthm}
\usepackage{array}
\usepackage{dirtytalk}
\usepackage{appendix}
\usepackage[dvipsnames]{xcolor}
\usepackage[font=small,labelfont=bf]{caption} 
\usepackage[nocompress]{cite}  
\usepackage{color,soul}  
\usepackage{dashrule}

\makeatletter
\DeclareRobustCommand\bfseries{%
  \not@math@alphabet\bfseries\mathbf
  \fontseries\bfdefault\selectfont
  \boldmath 
}
\makeatother

\newcommand{\myparagraph}[1]{
\vspace{0.1cm}\noindent
\textbf{#1.}
}

\newcommand{\myparagraphnp}[1]{
\vspace{0.1cm}\noindent
\textbf{#1}
}

\definecolor{orange}{rgb}{1,0.6,0.3}
\definecolor{dblue}{rgb}{0.80,0.85,0.9}
\definecolor{dorange}{RGB}{229,187,153}
\definecolor{lgreen}{RGB}{214,237,181}
\definecolor{lred}{RGB}{236,170,174}
\setulcolor{dorange}
\setul{0em}{0.15em}
\sethlcolor{dblue}

\DeclareRobustCommand{\hlg}[1]{{\sethlcolor{lgreen}\hl{#1}}}
\DeclareRobustCommand{\hlr}[1]{{\sethlcolor{lred}\hl{#1}}}

\newcommand{\data}{\mathcal{D}}

\newcommand{\loss}{\mathcal{L}}

\renewcommand{\vec}[1]{\bm{#1}}

\newcommand{\vecx}{\bm{x}}
\newcommand{\vecy}{\bm{y}}
\newcommand{\spacex}{\mathcal{X}}
\newcommand{\spacey}{\mathcal{Y}}

\newcommand{\simpi}{\stackrel{\pi}{\sim}}

\newcommand{\modelname}[1]{{\fontfamily{lmtt}\selectfont #1}}

\DeclareMathOperator*{\argmax}{arg\,max}

\usepackage[pagebackref=true,breaklinks=true,letterpaper=true,bookmarks=false]{hyperref}

\cvprfinalcopy 

\def\cvprPaperID{****} 

\ifcvprfinal\fi
\begin{document}

\title{Knockoff Nets: Stealing Functionality of Black-Box Models}

\author{
Tribhuvanesh Orekondy$^1$ \hspace{2cm} Bernt Schiele$^1$ \hspace{2cm} Mario Fritz$^2$ \vspace{0.5cm} \\
$^1$ Max Planck Institute for Informatics, Saarland Informatics Campus\\
$^2$ CISPA Helmholtz Center i.G., Saarland Informatics Campus
}

\maketitle

\begin{abstract}
Machine Learning (ML) models are increasingly deployed in the wild to perform a wide range of tasks.
In this work, we ask to what extent can an adversary steal functionality of such ``victim'' models based solely on blackbox interactions: image in, predictions out.
In contrast to prior work, we present an adversary lacking knowledge of train/test data used by the model, its internals, and semantics over model outputs.
We formulate model functionality stealing as a two-step approach: (i) querying a set of input images to the blackbox model to obtain predictions; and (ii) training a ``knockoff'' with queried image-prediction pairs.
We make multiple remarkable observations:
(a) querying random images from a different distribution than that of the blackbox training data results in a well-performing knockoff;
(b) this is possible even when the knockoff is represented using a different architecture; and
(c) our reinforcement learning approach additionally improves query sample efficiency in certain settings and provides performance gains.
We validate model functionality stealing on a range of datasets and tasks, as well as on a popular image analysis API where we create a reasonable knockoff for as little as \$30.
\end{abstract}

\section{Introduction}
\label{sec:introduction}
Machine Learning (ML) models and especially deep neural networks are deployed to improve productivity or experience e.g., photo assistants in smartphones, image recognition APIs in cloud-based internet services, and for navigation and control in autonomous vehicles.
Developing and engineering such models for commercial use is a product of intense time, money, and human effort -- ranging from collecting a massive annotated dataset to tuning the right model for the task.
The details of the dataset, exact model architecture, and hyperparameters are naturally kept confidential to protect the models' value.
However, in order to be monetized or simply serve a purpose, they are deployed in various applications (e.g., home assistants) to function as blackboxes: input in, predictions out.

Large-scale deployments of deep learning models in the wild has motivated the community to ask: can someone abuse the model solely based on blackbox access?
There has been a series of ``inference attacks'' \cite{shokri2017membership,joon2018iclr,fredrikson2015model,salem2018ml} which try to infer properties (e.g., training data \cite{shokri2017membership}, architecture \cite{joon2018iclr}) about the model within the blackbox.
In this work, we focus on model functionality stealing: can one create a ``knockoff'' of the blackbox model solely based on observed input-output pairs?
In contrast to previous works \cite{tramer2016stealing,papernot2017practical,joon2018iclr}, we work with minimal assumptions on the blackbox and intend to purely steal the \textit{functionality}.

\begin{figure}
    \begin{center}
       \includegraphics[width=\linewidth]{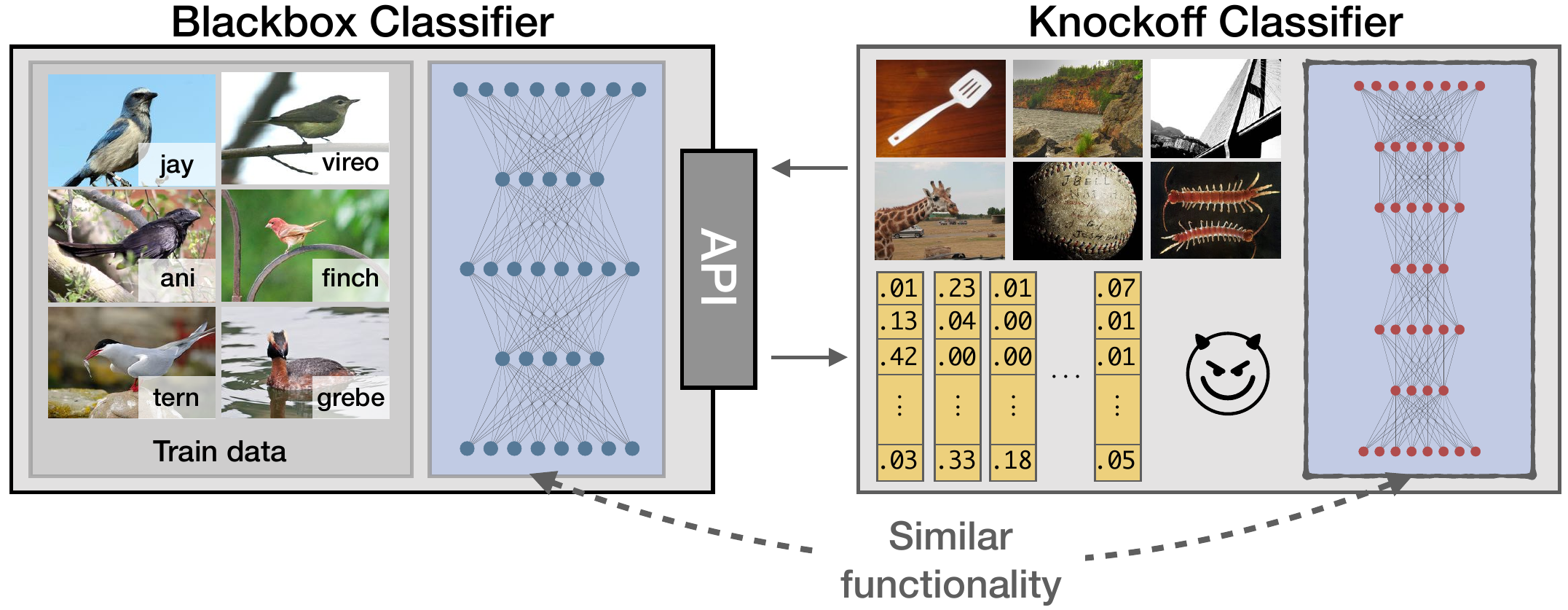}
    \end{center}
	\vspace{-1.2em}
   \caption{An adversary can create a ``knockoff'' of a blackbox model solely by interacting with its API: image in, prediction out.
   The knockoff bypasses the monetary costs and intellectual effort involved in creating the blackbox model.}
	\label{fig:teaser}
	\vspace{-1.0em}
\end{figure}

We formulate model functionality stealing as follows (shown in Figure~\ref{fig:teaser}).
The adversary interacts with a blackbox ``victim'' CNN by providing it input images and obtaining respective predictions.
The resulting image-prediction pairs are used to train a ``knockoff'' model.
The adversary's intention is for the knockoff to compete with the victim model at the victim's task.
Note that knowledge transfer \cite{hinton2015distilling,bucilua2006kdd} approaches are a special case within our formulation, where the task, train/test data, and white-box teacher (victim) model are known to the adversary.

Within this formulation, we spell out questions answered in our paper with an end-goal of model functionality stealing:
\begin{enumerate}[noitemsep,topsep=0pt,parsep=0pt,partopsep=0pt]
    \item Can we train a knockoff on a random set of query images and corresponding blackbox predictions?
    \item What makes for a good set of images to query?
    \item How can we improve sample efficiency of queries?
    \item What makes for a good knockoff architecture?
\end{enumerate}

\section{Related Work}
\label{sec:relatedwork}

\myparagraph{Privacy, Security and Computer Vision}
Privacy has been largely addressed within the computer vision community by proposing models \cite{orekondy2017towards,orekondy2017connecting,Wu_2018_ECCV,joon2016eccv,qianru2018cvpr,yonetani2017privacy} which recognize and control privacy-sensitive information in visual content.
The community has also recently studied security concerns entailing real-world usage of models e.g., adversarial perturbations \cite{moosavi2016deepfool,moosavi2017universal,joon2017iccv,akhtar2017defense,poursaeed2017generative,khrulkov2017art} in black- and white-box attack scenarios.
In this work, we focus on functionality stealing of CNNs in a blackbox attack scenario.

\myparagraph{Model Stealing}
Stealing various attributes of a blackbox ML model has been recently gaining popularity: parameters \cite{tramer2016stealing}, hyperparameters \cite{wang2018stealing}, architecture \cite{joon2018iclr}, information on training data \cite{shokri2017membership} and decision boundaries \cite{papernot2017practical}.
These works lay the groundwork to precisely reproduce the blackbox model.
In contrast, we investigate stealing \textit{functionality} of the blackbox independent of its internals.
Although two works \cite{tramer2016stealing,papernot2017practical} are related to our task, they make relatively stronger assumptions (e.g., model family is known, victim's data is partly available).
In contrast, we present a weaker adversary.

\myparagraph{Knowledge Distillation}
Distillation \cite{hinton2015distilling} and related approaches \cite{furlanello2018born, ying2018DML,chen2017learning,bucilua2006kdd} transfer the knowledge from a complex ``teacher'' to a simpler ``student'' model.
Within our problem formulation, this is a special case when the adversary has strong knowledge of the victim's blackbox model e.g., architecture, train/test data is known.
Although we discuss this, a majority of the paper makes weak assumptions of the blackbox.

\myparagraph{Active Learning}
Active Learning \cite{tong2001support,cohn1996active} (AL) aims to reduce labeling effort while gathering data to train a model.
Ours is a special case of pool-based AL \cite{settles2008analysis}, where the learner (adversary) chooses from a pool of unlabeled data.
However, unlike AL, the learner's image pool in our case is chosen without any knowledge of the data used by the original model.
Moreover, while AL considers the image to be annotated by a human-expert, ours is annotated with pseudo-labels by the blackbox.

\section{Problem Statement}
\label{sec:problem}

\begin{figure}
    \begin{center}
       \includegraphics[width=\linewidth]{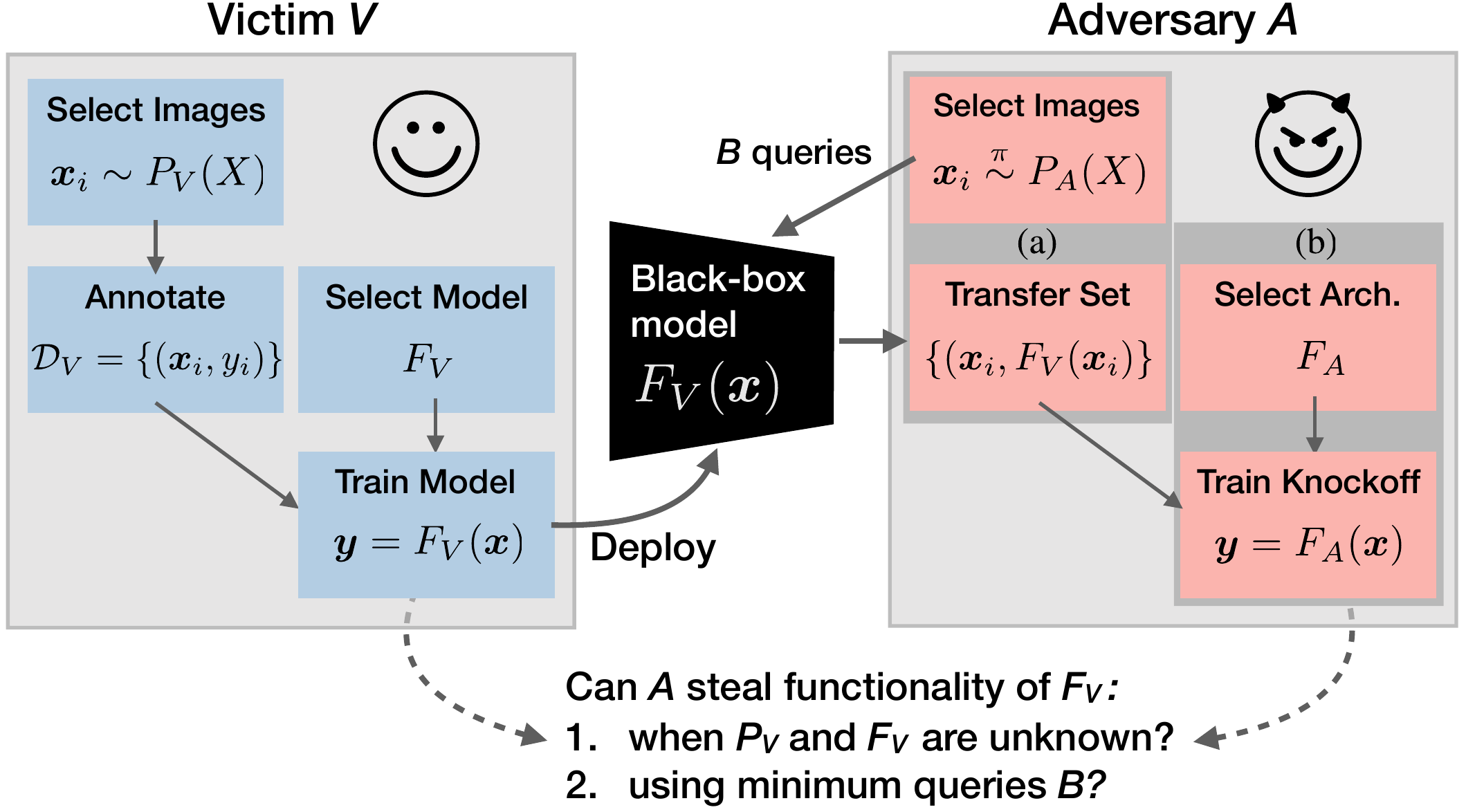}
    \end{center}
	\vspace{-1.0em}
   \caption{\textbf{Problem Statement.} Laying out the task of model functionality stealing in the view of two players - victim $V$ and adversary $A$. We group adversary's moves into (a) Transfer Set Construction (b) Training Knockoff $F_A$.}
	\label{fig:problem_statement}
	\vspace{-1.0em}
\end{figure}

We now formalize the task of functionality stealing (see also Figure~\ref{fig:problem_statement}).

\myparagraph{Functionality Stealing}
In this paper, we introduce the task as:
given blackbox query access to a ``victim'' model $F_V: \spacex \rightarrow \spacey$, to replicate its functionality using ``knockoff'' model $F_A$ of the adversary.
As shown in Figure~\ref{fig:problem_statement}, we set it up as a two-player game between a victim $V$ and an adversary $A$.
Now, we discuss the assumptions in which the players operate and their corresponding moves in this game.

\myparagraph{Victim's Move}
The victim's end-goal is to deploy a trained CNN model $F_V$ in the wild for a particular task (e.g., fine-grained bird classification).
To train this particular model, the victim:
(i) collects task-specific images $\vecx \sim P_V(X)$ and obtains expert annotations resulting in a dataset $\data_V = \{(\vecx_i, y_i)\}$;
(ii) selects the model $F_V$ that achieves best performance (accuracy) on a held-out test set of images $\data_V^\text{test}$.
The resulting model is deployed as a blackbox which predicts output probabilities $\vecy = F_V(\vecx)$ given an image $\vecx$.
Furthermore, we assume each prediction incurs a cost (e.g., monetary, latency).

\myparagraph{Adversary's Unknowns}
The adversary is presented with a blackbox CNN image classifier, which given \textit{any} image $\vecx \in \spacex$ returns a $K$-dim posterior probability vector $\vecy \in [0, 1]^K,\ \sum_k y_k = 1$.
We relax this later by considering truncated versions of $\vecy$.
We assume remaining aspects to be unknown: 
(i) the internals of $F_V$ e.g., hyperparameters or architecture;
(ii) the data used to train and evaluate the model; and
(iii) semantics over the $K$ classes.

\myparagraph{Adversary's Attack}
To train a knockoff, the adversary:
(i) interactively queries images $\{\vecx_i \simpi P_A(X)\}$ using strategy $\pi$ to obtain a ``transfer set'' of images and pseudo-labels $\{(\vecx_i, F_V(\vecx_i))\}_{i=1}^B$; and
(ii) selects an architecture $F_A$ for the knockoff and trains it to mimic the behaviour of $F_V$ on the transfer set.

\myparagraph{Objective}
We focus on the adversary, whose primary objective is training a knockoff that performs well on the task for which $F_V$ was designed i.e., on an unknown $\data_V^\text{test}$.
In addition, we address two secondary objectives:
(i) sample-efficiency: maximizing performance within a budget of $B$ blackbox queries; and
(ii) understanding what makes for good images to query the blackbox.

\myparagraph{Victim's Defense}
Although we primarily address the adversary's strategy in the paper, we briefly discuss victim's counter strategies (in Section \ref{sec:results}) of reducing informativeness of predictions by truncation e.g., rounding-off.

\begin{figure}
    \begin{center}
       \includegraphics[width=\linewidth]{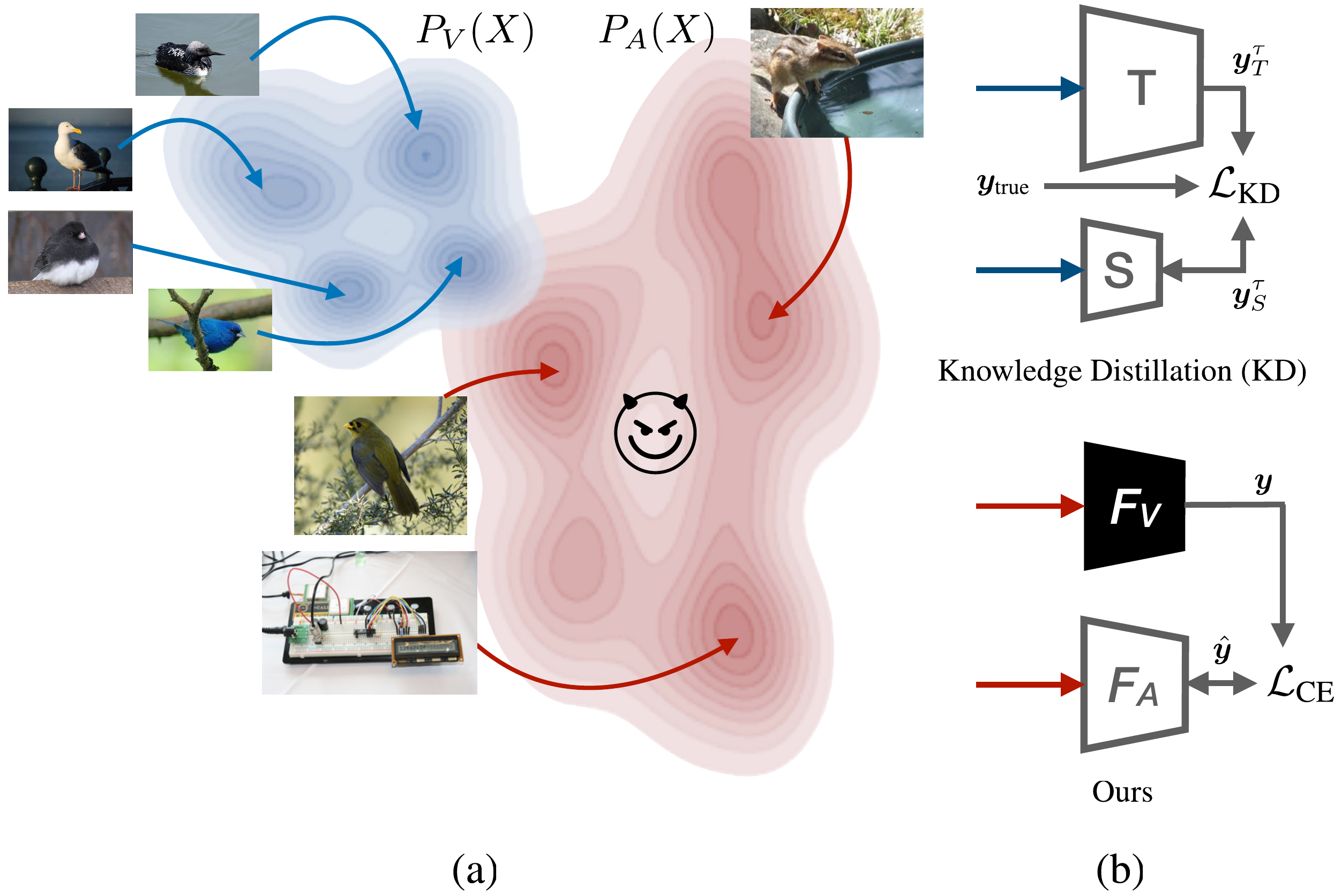}
    \end{center}
	\vspace{-1.0em}
   \caption{\textbf{Comparison to KD.} (a) Adversary has access only to image distribution $P_A(X)$ (b) Training in a KD-manner requires stronger knowledge of the victim. Both $S$ and $F_A$ are trained to classify images $\vecx \in P_V(X)$}
	\label{fig:imagesets}
	\vspace{-1.0em}
\end{figure}

\myparagraph{Remarks: Comparison to Knowledge Distillation (KD)}
Training the knockoff model is  reminiscent of KD approaches \cite{hinton2015distilling,romero2014fitnets}, whose goal is to transfer the knowledge from a larger teacher network $T$ (white-box) to a compact student network $S$ (knockoff) via the transfer set.
We illustrate key differences between KD and our setting in Figure~\ref{fig:imagesets}:
(a) \textbf{Independent distribution $P_A$}: $F_A$ is trained on images $\vecx \sim P_A(X)$ \textit{independent} to distribution $P_V$ used for training $F_V$;
(b) \textbf{Data for supervision}: Student network $S$ minimize variants of KD loss:
\begin{equation*}
    \loss_{\text{KD}} = \lambda_1 \loss_{\text{CE}}(\vecy_{\text{true}},\ \vecy_S) + \lambda_2 \loss_{\text{CE}}(\vecy_S^\tau,\ \vecy_T^\tau)
\end{equation*}
where $\vecy^\tau_T = \text{softmax}(\vec{a}_T/\tau)$ is the softened posterior distribution of logits $\vec{a}$ controlled by temperature $\tau$.
In contrast, the knockoff (student) in our case lacks logits $\vec{a}_T$ and true labels $\vecy_{\text{true}}$ to supervise training.

\section{Generating Knockoffs}
\label{sec:approach}
In this section, we elaborate on the adversary's approach in two steps:
transfer set construction (Section \ref{sec:approach_query}) and training knockoff $F_A$ (Section \ref{sec:approach_fa}).

\subsection{Transfer Set Construction}
\label{sec:approach_query}
The goal is to obtain a transfer set i.e., image-prediction pairs, on which the knockoff will be trained to imitate the victim's blackbox model $F_V$.

\myparagraph{Selecting $P_A(X)$}
The adversary first selects an image distribution to sample images.
We consider this to be a large discrete set of images.
For instance, one of the distributions $P_A$ we consider is the 1.2M images of ILSVRC dataset \cite{deng2009imagenet}.

\myparagraph{Sampling Strategy $\pi$}
Once the image distribution $P_A(X)$ is chosen, the adversary samples images $\vecx \simpi P_A(X)$ using a strategy $\pi$.
We consider two strategies.

\begin{figure}
    \begin{center}
       \includegraphics[width=0.9\linewidth]{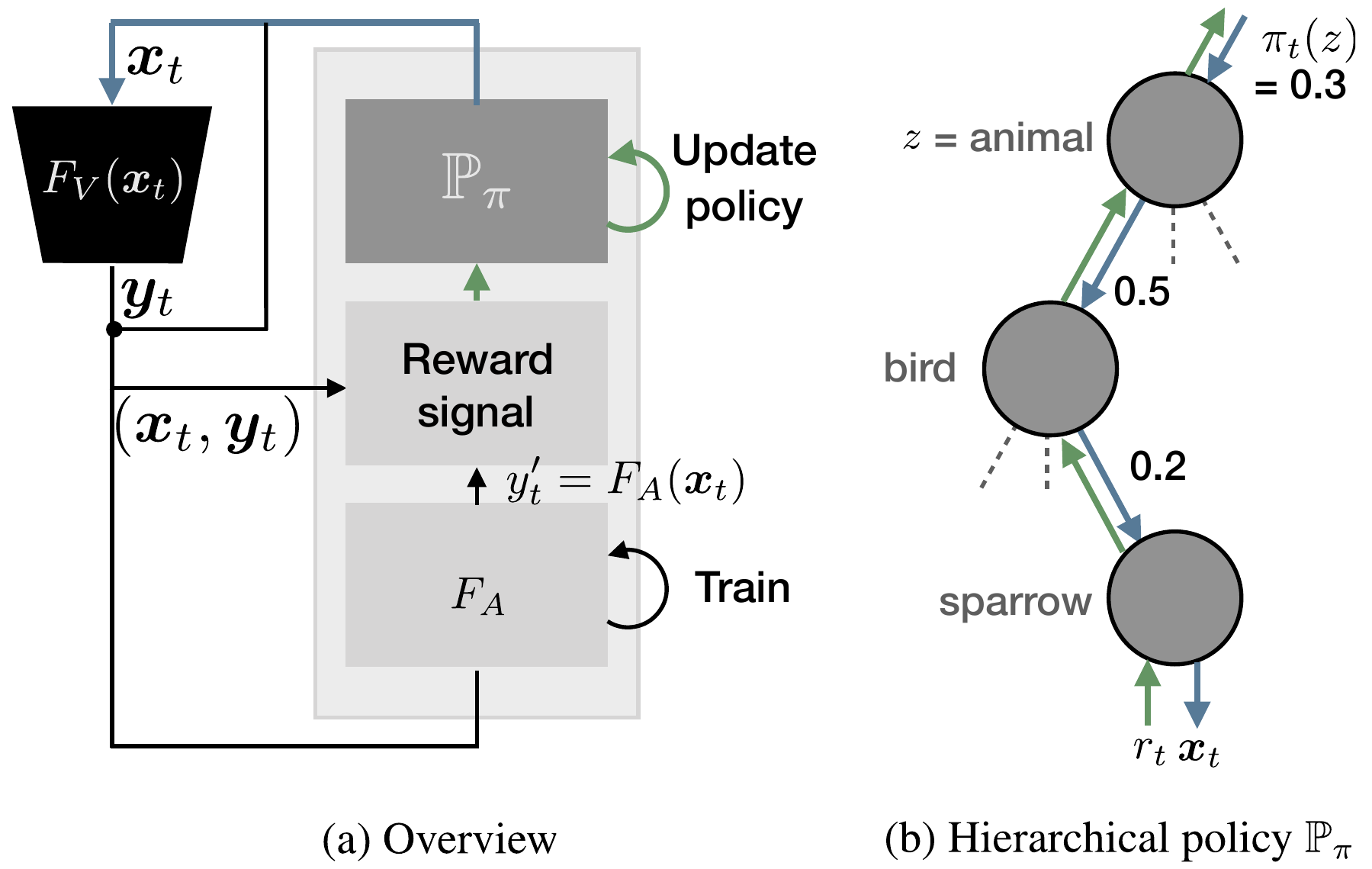}
    \end{center}
	\vspace{-1.0em}
   \caption{\textbf{Strategy \texttt{adaptive}.}}
	\label{fig:agent}
	\vspace{-1.0em}
\end{figure}

\subsubsection{\texttt{Random} Strategy}
\label{sec:approach_query_random}
In this strategy, we randomly sample images (without replacement) $\vecx \stackrel{\text{iid}}{\sim} P_A(X)$ to query $F_V$.
This is an extreme case where adversary performs pure exploration.
However, there is a risk that the adversary samples images irrelevant to learning the task (e.g., over-querying dog images to a birds classifier).

\subsubsection{\texttt{Adaptive} Strategy}
\label{sec:approach_query_adaptive}
We now incorporate a feedback signal resulting from each image queried to the blackbox.
A policy $\pi$ is learnt:
\begin{equation*}
    \vecx_t \sim \mathbb{P}_{\pi}( \{\vecx_i, \vecy_i\}_{i=1}^{t-1} )
\end{equation*}
to achieve two goals:
(i) improving sample-efficiency of queries; and 
(ii) aiding interpretability of blackbox $F_V$.
The approach is outlined in Figure~\ref{fig:agent}a.
At each time-step $t$, the policy module $\mathbb{P}_{\pi}$ produces a sample of images to query the blackbox.
A reward signal $r_t$ is shaped based on multiple criteria and is used to update the policy with an end-goal of maximizing the expected reward.

\myparagraph{Supplementing $\bm{P_A}$}
To encourage relevant queries, we enrich images in the adversary's distribution by associating each image $\vecx_i$ with a label $z_i \in Z$.
No semantic relation of these labels with the blackbox's output classes is assumed or exploited.
As an example, when $P_A$ corresponds to 1.2M images of the ILSVRC \cite{deng2009imagenet} dataset, we use labels defined over 1000 classes.
These labels can be alternatively obtained by unsupervised measures e.g., clustering or estimating graph-density \cite{ebert2012ralf,beluch2018power}.
We find using labels aids understanding blackbox functionality.
Furthermore, since we expect labels $\{z_i \in Z\}$ to be correlated or inter-dependent, we represent them within a coarse-to-fine hierarchy, as nodes of a tree as shown in Figure~\ref{fig:agent}b.

\myparagraph{Actions}
At each time-step $t$, we sample actions from a discrete action space $z_t \in Z$ i.e., adversary's independent label space.
Drawing an action is a forward-pass (denoted by a blue line in Figure~\ref{fig:agent}b) through the tree: at each level, we sample a node with probability $\pi_t(z)$ .
The probabilities are determined by a softmax distribution over the node potentials: $\pi_t(z) = \frac{e^{H_t(z)}}{\sum_{z'}H_t(z')}$.
Upon reaching a leaf-node, a sample of images is returned corresponding to label $z_t$.

\myparagraph{Learning the Policy}
We use the received reward $r_t$ for an action $z_t$ to update the policy $\pi$ using the gradient bandit algorithm \cite{sutton1998introduction}.
This update is equivalent to a backward-pass through the tree (denoted by a green line in Figure~\ref{fig:agent}b), where the node potentials are updated as:
\begin{align*} 
    H_{t+1}(z_t) &= H_t(z_t) + \alpha (r_t - \bar{r}_t) (1 - \pi_t(z_t)) & \text{and} \\
    H_{t+1}(z')   &= H_t(z') + \alpha (r_t - \bar{r}_t)\pi_t(z')   & \forall z' \ne z_t
\end{align*}
where $\alpha = 1/N(z)$ is the learning rate, $N(z)$ is the number of times action $z$ has been drawn, and $\bar{r}_t$ is the mean-reward over past $\Delta$ time-steps.

\myparagraph{Rewards}
To evaluate the quality of sampled images $\vecx_t$, we study three rewards.
We use a margin-based \textbf{certainty} measure \cite{settles2008analysis,joshi2009multi} to encourage images where the victim is confident (hence indicating the domain $F_V$ was trained on):
\begin{equation}
    \vspace{-0.5em}
    \tag*{``\texttt{cert}''}
    R^{\text{cert}}(\vecy_t) = P(\vecy_{t, k_1} | \vecx_t) - P(\vecy_{t, k_2} | \vecx_t)
\end{equation}
To prevent the degenerate case of image exploitation over a single label, we introduce a \textbf{diversity} reward:
\begin{equation}
    \vspace{-0.5em}
    \tag*{``\texttt{div}''}
    R^{\text{div}}(\vecy_{1:t}) = \sum_k \max(0, \bar{\vecy}_{t, k} - \bar{\vecy}_{t-\Delta, k})
    \label{eqn:reward_div}
\end{equation}
To encourage images where the knockoff prediction $\hat{\vecy}_t = F_A(\vecx_t)$ does not imitate $F_V$, we reward high \textbf{loss}:
\begin{equation}
    \vspace{-0.5em}
    \tag*{\text{``$\loss$''}}
    R^{\loss}(\vecy_t, \hat{\vecy}_t) = \mathcal{L}(\vecy_t, \hat{\vecy_t})
    \label{eqn:reward_loss}
\end{equation}
We sum up individual rewards when multiple measures are used.
To maintain an equal weighting, each reward is individually rescaled to [0, 1] and subtracted with a baseline computed over past $\Delta$ time-steps.

\subsection{Training Knockoff $F_A$}
\label{sec:approach_fa}
As a product of the previous step of interactively querying the blackbox model, we have a transfer set $\{(\vecx_t, F_V(\vecx_t)\}_{t=1}^B,\ \vecx_t \simpi P_A(X)$.
Now we address how this is used to train a knockoff $F_A$.

\myparagraph{Selecting Architecture $F_A$}
Few works \cite{joon2018iclr,wang2018stealing} have recently explored reverse-engineering the blackbox i.e., identifying the architecture, hyperparameters, etc.
We however argue this is orthogonal to our requirement of simply stealing the functionality.
Instead, we represent $F_A$ with a reasonably complex architecture e.g., VGG \cite{simonyan2014very} or ResNet \cite{he2016deep}.
Existing findings in KD \cite{hinton2015distilling,furlanello2018born} and model compression \cite{bucilua2006kdd,han2015deep,iandola2016squeezenet} indicate robustness to choice of reasonably complex student models.
We investigate the choice under weaker knowledge of the teacher ($F_V$) e.g., training data and architecture is unknown.

\myparagraph{Training to Imitate}
To bootstrap learning, we begin with a pretrained Imagenet network $F_A$.
We train the knockoff $F_A$ to imitate $F_V$ on the transfer set 
by minimizing the cross-entropy (CE) loss:
 $\mathcal{L}_{\text{CE}}(\vecy, \hat{\vecy}) = - \sum_k p(y_k) \cdot \log p(\hat{y_k})$.
This is a standard CE loss, albeit weighed with the confidence $p(y_k)$ of the victim's label.
This formulation is equivalent to minimizing the KL-divergence between the victim's and knockoff's predictions over the transfer set.

\section{Experimental Setup}
\label{sec:exptsetup}
We now discuss the experimental setup of multiple victim blackboxes (Section \ref{sec:exptsetup_fv}), followed by details on the adversary's approach (Section \ref{sec:exptsetup_pa}).

\subsection{Black-box Victim Models $F_V$}
\label{sec:exptsetup_fv}
We choose four diverse image classification CNNs, addressing multiple challenges in image classification e.g., fine-grained recognition.
Each CNN performs a task specific to a dataset.
A summary of the blackboxes is presented in Table~\ref{tab:blackboxes} (extended descriptions in appendix).

\begin{table}[]
    \centering
    \footnotesize
    \begin{tabular}{@{}lll@{}}
    \toprule
    Blackbox ($F_V$)         & $|\data_V^{\text{train}}| + |\data_V^{\text{test}}|$     & Output classes $K$                             \\ \midrule
    \modelname{Caltech256} \cite{griffin2007caltech} & 23.3k + 6.4k   & 256 general object categories \\
    \modelname{CUBS200} \cite{WahCUB_200_2011}    & 6k + 5.8k  & 200 bird species              \\
    \modelname{Indoor67} \cite{quattoni2009recognizing}   & 14.3k + 1.3k & 67 indoor scenes              \\
    \modelname{Diabetic5} \cite{eyepacs}  & 34.1k + 1k   & 5 diabetic retinopathy scales    \\ \bottomrule
    \end{tabular}
    \caption{\textbf{Four victim blackboxes $F_V$}. Each blackbox is named in the format: [dataset][\# output classes].}
    \label{tab:blackboxes}
	\vspace{-1.5em}
\end{table}

\myparagraph{Training the Black-boxes}
All models are trained using a ResNet-34 architecture (with ImageNet \cite{deng2009imagenet} pretrained weights) on the training split of the respective datasets.
We find this architecture choice achieve strong performance on all datasets at a reasonable computational cost.
Models are trained using SGD with momentum (of 0.5) optimizer for 200 epochs with a base learning rate of 0.1 decayed by a factor of 0.1 every 60 epochs.
We follow the train-test splits suggested by the respective authors for \textbf{Caltech-256} \cite{griffin2007caltech}, \textbf{CUBS-200-2011} \cite{WahCUB_200_2011}, and \textbf{Indoor-Scenes} \cite{quattoni2009recognizing}.
Since GT annotations for \textbf{Diabetic-Retinopathy} \cite{eyepacs} test images are not provided, 
we reserve 200 training images for each of the five classes for testing.
The number of test images per class for all datasets are roughly balanced.
The test images of these datasets $\data_V^{\text{test}}$ are used to evaluate both the victim and knockoff models.

After these four victim models are trained, we use them as a  blackbox for the remainder of the paper: images in, posterior probabilities out.

\subsection{Representing $P_A$}
\label{sec:exptsetup_pa}

In this section, we elaborate on the setup of two aspects relevant to transfer set construction (Section \ref{sec:approach_query}).

\subsubsection{Choice of $P_A$}
\label{sec:exptsetup_pa_choice}
Our approach for transfer set construction involves the adversary querying images from a large discrete image distribution $P_A$.
In this section, we present four choices considered in our experiments.
Any information apart from the images from the respective datasets are unused in the \texttt{random} strategy.
For the \texttt{adaptive} strategy, we use image-level labels (chosen independent of blackbox models) to guide sampling.

\myparagraph{$\bm{P_A = P_V}$}
For reference, we sample from the exact set of images used to train the blackboxes.
This is a special case of knowledge-distillation \cite{hinton2015distilling} with unlabeled data at temperature $\tau=1$.

\myparagraph{$\bm{P_A = }$ ILSVRC \cite{russakovsky2015imagenet,deng2009imagenet}}
We use the collection of 1.2M images over 1000 categories  presented in the ILSVRC-2012 \cite{russakovsky2015imagenet} challenge.

\myparagraph{$\bm{P_A = }$ OpenImages \cite{kuznetsova2018open}}
OpenImages v4 is a large-scale dataset of 9.2M images gathered from Flickr.
We use a subset of 550K unique images, gathered by sampling 2k images from each of 600 categories.

\myparagraph{$\bm{P_A = D^2}$}
We construct a dataset wherein the adversary has access to all images in the universe.
In our case, we create the dataset by pooling training data from:
(i) all four datasets listed in Section \ref{sec:exptsetup_fv}; and
(ii) both datasets presented in this section.
This results in a ``dataset of datasets'' $D^2$ of 2.2M images and 2129 classes.

\myparagraph{Overlap between $P_A$ and $P_V$}
We compute overlap between labels of the blackbox ($K$, e.g., 256 Caltech classes) and the adversary's dataset ($Z$, e.g., 1k ILSVRC classes) as: $100 \times |K \cap Z| / |K|$.
Based on the overlap between the two image distributions, we categorize $P_A$ as:
\begin{enumerate}[noitemsep,topsep=0pt,parsep=0pt,partopsep=0pt]
    \item $\bm{P_A = P_V}$: Images queried are identical to the ones used for training $F_V$. There is a 100\% overlap.
    \item \textbf{Closed-world ($\bm{P_A = D^2}$)}: Blackbox train data $P_V$ is a subset of the image universe $P_A$. There is a 100\% overlap.
    \item \textbf{Open-world ($\bm{P_A \in }$ \{ILSVRC, OpenImages\})}: Any overlap between $P_V$ and $P_A$ is purely coincidental. Overlaps are: \modelname{Caltech256} (42\% ILSVRC, 44\% OpenImages), \modelname{CUBS200} (1\%, 0.5\%), \modelname{Indoor67} (15\%, 6\%), and \modelname{Diabetic5} (0\%, 0\%). 
\end{enumerate}

\subsubsection{\texttt{Adaptive} Strategy}
In the adaptive strategy (Section \ref{sec:approach_query_adaptive}), we make use of auxiliary information (labels) in the adversary's data $P_A$ to guide the construction of the transfer set.
We represent these labels as the leaf nodes in the coarse-to-fine concept hierarchy tree.
The root node in all cases is a single concept ``entity''.
We obtain the rest of the hierarchy as follows:
(i) $D^2$: we add as parents the dataset the images belong to;
(ii) ILSVRC: for each of the 1K labels, we obtain 30 coarse labels by clustering the mean visual features of each label obtained using 2048-dim pool features of an ILSVRC pretrained Resnet model;
(iii) OpenImages: We use the exact hierarchy provided by the authors.

\section{Results}
\label{sec:results}
We now discuss the experimental results.

\myparagraph{Training Phases}
The knockoff models are trained in two phases:
(a) \textit{Online}: during transfer set construction (Section \ref{sec:approach_query}); followed by
(b) \textit{Offline}: the model is retrained using transfer set obtained thus far (Section \ref{sec:approach_fa}).
All results on knockoff are reported after step (b).

\myparagraph{Evaluation Metric}
We evaluate two aspects of the knockoff: (a) \textit{Top-1 accuracy}: computed on victim's held-out test data $\data_V^{\text{test}}$ (b) \textit{sample-efficiency}: best performance achieved after a budget of $B$ queries.
Accuracy is reported in two forms: absolute ($x$\%) or relative to blackbox $F_V$ ($x\times$).

\begin{table*}[t]
\centering
\scriptsize
\begin{tabular}{@{}lllllllllll@{}}
\toprule
                      &                      & \multicolumn{4}{c}{\texttt{random}}                                                            &  & \multicolumn{4}{c}{\texttt{adaptive}}                                                          \\ \cmidrule{3-6} \cmidrule{8-11}
                      & $P_A$                      & \modelname{Caltech256}          & \modelname{CUBS200}                & \modelname{Indoor67}              & \modelname{Diabetic5}            &  & \modelname{Caltech256}          & \modelname{CUBS200}                & \modelname{Indoor67}              & \modelname{Diabetic5}            \\ \midrule 
                      & \color{gray}{$P_V (F_V)$}                 & \color{gray}{78.8 (1$\times$)}    & \color{gray}{76.5 (1$\times$)}    & \color{gray}{74.9 (1$\times$)}    & \color{gray}{58.1 (1$\times$)}    &  & -                   & -                   & -                   & -                   \\
                      & $P_V$ (KD)                 & 82.6 (1.05$\times$) & 70.3 (0.92$\times$) & 74.4 (0.99$\times$) & 54.3 (0.93$\times$) &  & -                   & -                   & -                   & -                   \\ \cmidrule{2-11}
Closed                & $D^2$ & 76.6 (0.97$\times$) & 68.3 (0.89$\times$) & 68.3 (0.91$\times$) & 48.9 (0.84$\times$) &  & 82.7 (1.05$\times$) & 74.7 (0.98$\times$) & 76.3 (1.02$\times$) & 48.3 (0.83$\times$) \\  \cmidrule{2-11}
\multirow{2}{*}{Open} & ILSVRC               & 75.4 (0.96$\times$) & 68.0 (0.89$\times$)   & 66.5 (0.89$\times$) & 47.7 (0.82$\times$) &  & 76.2 (0.97$\times$) & 69.7 (0.91$\times$) & 69.9 (0.93$\times$) & 44.6 (0.77$\times$) \\
                      & OpenImg              & 73.6 (0.93$\times$) & 65.6 (0.86$\times$) & 69.9 (0.93$\times$) & 47.0 (0.81$\times$)   &  & 74.2 (0.94$\times$) & 70.1 (0.92$\times$) & 70.2 (0.94$\times$) & 47.7 (0.82$\times$) \\ \bottomrule
\end{tabular}
\caption{\textbf{Accuracy on test sets.} 
Accuracy of {\color{gray}{blackbox $F_V$}} indicated in gray and knockoffs $F_A$ in black.
KD = Knowledge Distillation.
Closed- and open-world accuracies reported at $B$=60k.
}
\label{tab:pa_summary}
\end{table*}

\begin{figure*}
    \begin{center}
       \includegraphics[width=\linewidth]{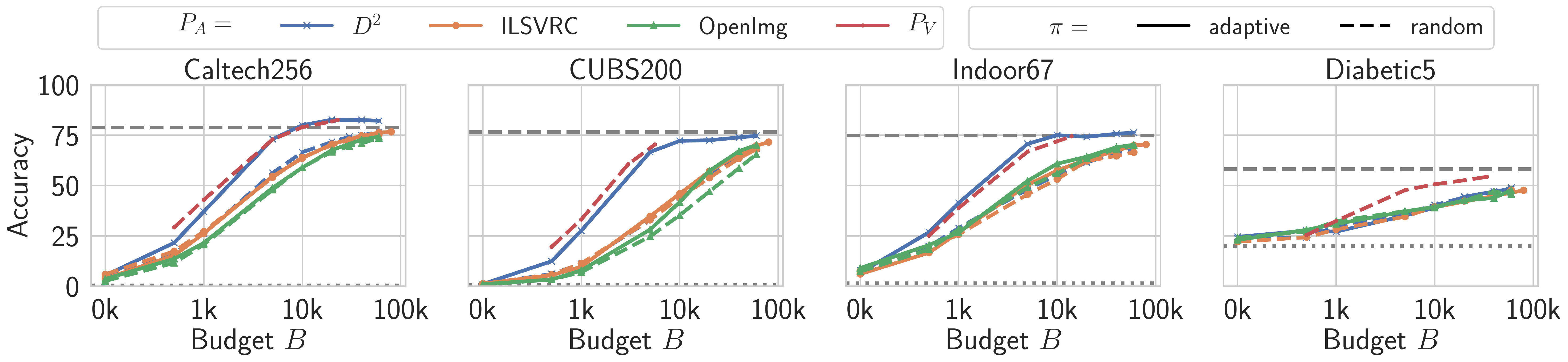}
    \end{center}
	\vspace{-1.5em}
   \caption{\textbf{Performance of the knockoff at various budgets.} Presented for various choices of adversary's image distribution ($P_A$) and sampling strategy $\pi$. 
   {\color{gray}{\hdashrule[0.5ex]{0.3cm}{1pt}{1mm 1.2pt}}} represents accuracy of blackbox $F_V$ and {\color{gray}{\hdashrule[0.5ex]{0.4cm}{1pt}{1pt}}} represents chance-level performance.}
	\label{fig:expt_pa}
	\vspace{-1.0em}
\end{figure*}

In each of the following experiments, we evaluate our approach with identical hyperparameters across all blackboxes, highlighting the generalizability of model functionality stealing.

\subsection{Transfer Set Construction}
\label{sec:results_transfer}
In this section, we analyze influence of transfer set $\{(\vecx_i, F_V(\vecx_i)\}$ on the knockoff.
For simplicity, for the remainder of this section we fix the architecture of the victim and knockoff to a Resnet-34 \cite{he2016deep}.

\myparagraph{Reference: $\bm{P_A = P_V}$ (KD)}
From Table~\ref{tab:pa_summary} (second row), we observe:
(i) all knockoff models recover $0.92$-$1.05\times$ performance of $F_V$;
(ii) a better performance than $F_V$ itself (e.g., 3.8\% improvement on \modelname{Caltech256}) due to regularizing effect of training on soft-labels \cite{hinton2015distilling}.


\begin{figure*}
    \begin{center}
       \includegraphics[width=\linewidth]{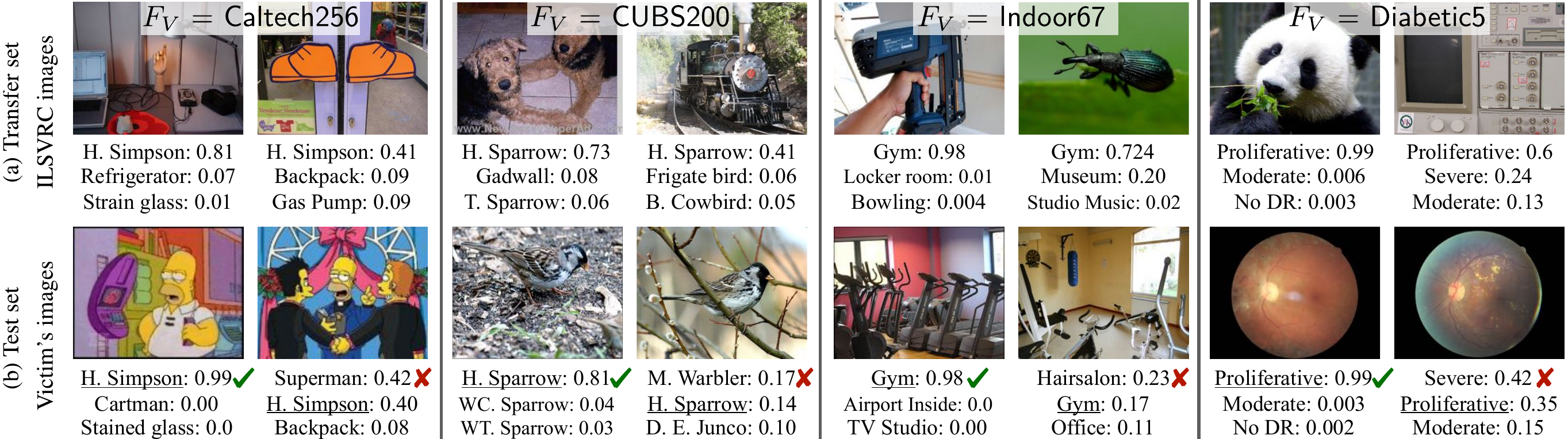}
    \end{center}
	\vspace{-1.0em}
   \caption{\textbf{Qualitative Results. } (a) Samples from the transfer set ($\{ (\vecx_i, F_V(\vecx_i)) \}, \vecx_i \sim P_A(X)$) displayed for four output classes (one from each blackbox): `Homer Simpson', `Harris Sparrow', `Gym', and `Proliferative DR'.
   (b) With the knockoff $F_A$ trained on the transfer set, we visualize its predictions on victim's test set ($\{ (\vecx_i, F_A(\vecx_i)) \}, \vecx_i \sim \data_V^{\text{test}}$). 
   \underline{Ground truth} labels are underlined.
   Objects from these classes, among numerous others, were never encountered while training $F_A$.}
	\label{fig:expt_qual}
	\vspace{-1.0em}
\end{figure*}

\myparagraphnp{Can we learn by querying \textit{randomly} from an independent distribution?}
Unlike KD, the knockoff is now trained and evaluated on different image distributions ($P_A$ and $P_V$ respectively).
We first focus on the \texttt{random} strategy, which does not use any auxiliary information.

We make the following observations from Table~\ref{tab:pa_summary} (\texttt{random}):
(i) \textbf{closed-world}: 
the knockoff is able to reasonably imitate all the blackbox models, recovering $0.84$-$0.97\times$ blackbox performance;
(ii) \textbf{open-world}: 
in this challenging scenario, the knockoff model has \textit{never} encountered images of numerous classes at test-time e.g., $>$90\% of the bird classes in \modelname{CUBS200}.
Yet remarkably, the knockoff is able to obtain $0.81$-$0.96\times$ performance of the blackbox.
Moreover, results marginally vary (at most $0.04\times$) between ILSVRC and OpenImages, indicating any large diverse set of images makes for a good transfer set.

Upon qualitative analysis, we find the image and pseudo-label pairs in the transfer set are semantically incoherent (Fig.~\ref{fig:expt_qual}a) for output classes non-existent in training images $P_A$.
However, when relevant images are presented at test-time (Fig.~\ref{fig:expt_qual}b), the adversary displays strong performance.
Furthermore, we find the top predictions by knockoff relevant to the image e.g., predicting one comic character (superman) for another.

\myparagraphnp{How \textit{sample-efficient} can we get?}
Now we evaluate the \texttt{adaptive} strategy (discussed in Section \ref{sec:approach_query_adaptive}).
Note that we make use of auxiliary information of the images in these tasks (labels of images in $P_A$).
We use the reward set which obtained the best performance in each scenario: \{certainty\} in closed-world and \{certainty, diversity, loss\} in open-world.

From Figure~\ref{fig:expt_pa}, we observe:
(i) \textbf{closed-world}:
\texttt{adaptive} is extremely sample-efficient in all but one case.
Its performance is comparable to KD in spite of samples drawn from a $36$-$188\times$ larger image distribution.
We find significant sample-efficiency improvements e.g., while \texttt{CUBS200-random} reaches 68.3\% at $B$=60k, \texttt{adaptive} achieves this 6$\times$ quicker at $B$=10k.
We find comparably low performance in \modelname{Diabetic5} as the blackbox exhibits confident predictions for all images resulting in poor feedback signal to guide policy;
(ii) \textbf{open-world}:
although we find marginal improvements over \texttt{random} in this challenging scenario, they are pronounced in few cases e.g., $1.5\times$ quicker to reach an accuracy 57\% on \texttt{CUBS200} with OpenImages.
(iii) as an added-benefit apart from sample-efficiency, from Table~\ref{tab:pa_summary}, we find \texttt{adaptive} display improved performance (up to 4.5\%) consistently across all choices of $F_V$.

\begin{figure}
    \begin{center}
       \includegraphics[width=\linewidth]{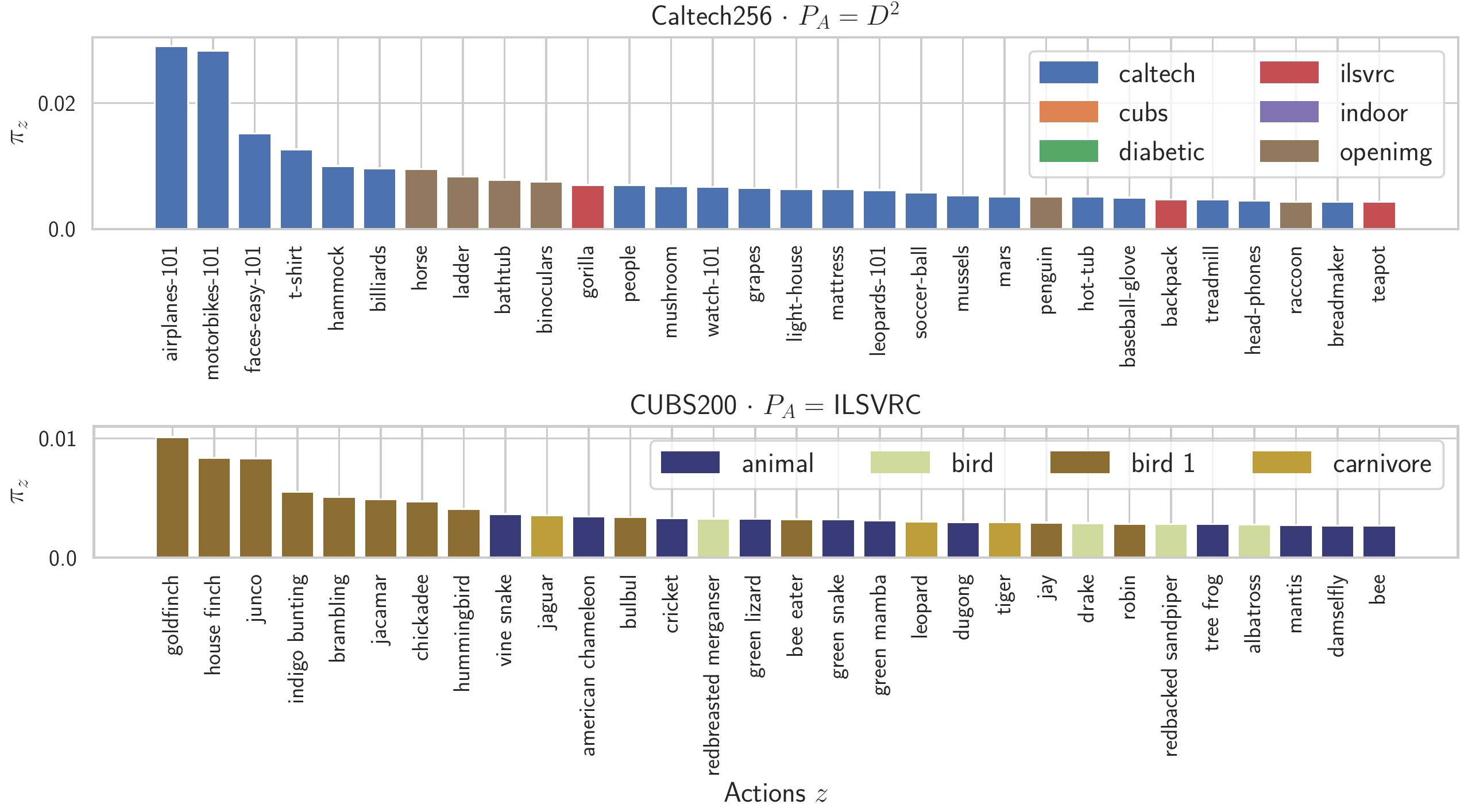}
    \end{center}
	\vspace{-1.0em}
   \caption{\textbf{Policy $\pi$ learnt by the \texttt{adaptive} approach.} Each bar represents preference for action $z$. Top 30 actions (out of 2.1k and 1k) are displayed. Colors indicate parent of action in hierarchy.}
	\label{fig:expt_policy}
	\vspace{-0.5em}
\end{figure}

\myparagraphnp{What can we learn by inspecting the \textit{policy}?}
From previous experiments, we observed two benefits of the \texttt{adaptive} strategy: sample-efficiency (although more prominent in the closed-world) and improved performance.
The policy $\pi_t$ learnt by \texttt{adaptive} (Section \ref{sec:approach_query_adaptive}) additionally allows us to understand what makes for good images to query.
$\pi_t(z)$ is a discrete probability distribution indicating preference over action $z$. 
Each action $z$ in our case corresponds to labels in the adversary's image distribution.

We visualize $\pi_t(z)$ in Figure~\ref{fig:expt_policy}, where each bar represents an action and its color, the parent in the hierarchy.
We observe:
(i) \textbf{closed-world} (Fig.~\ref{fig:expt_policy} top):
actions sampled with higher probabilities consistently correspond to output classes of $F_V$.
Upon analyzing parents of these actions (the dataset source), the policy also learns to sample images for the output classes from an alternative richer image source e.g., ``ladder'' images in \modelname{Caltech256} sampled from OpenImages instead;
(ii) \textbf{open-world} (Fig.~\ref{fig:expt_policy} bottom):
unlike closed-world, the optimal mapping between adversary's actions to blackbox's output classes is non-trivial and unclear.
However, we find top actions typically correspond to output classes of $F_V$ e.g., indigo bunting.
The policy, in addition, learns to sample coarser actions related to the $F_V$'s task e.g., predominantly drawing from birds and animals images to knockoff \modelname{CUBS200}.

\begin{figure}
    \begin{center}
       \includegraphics[width=\linewidth]{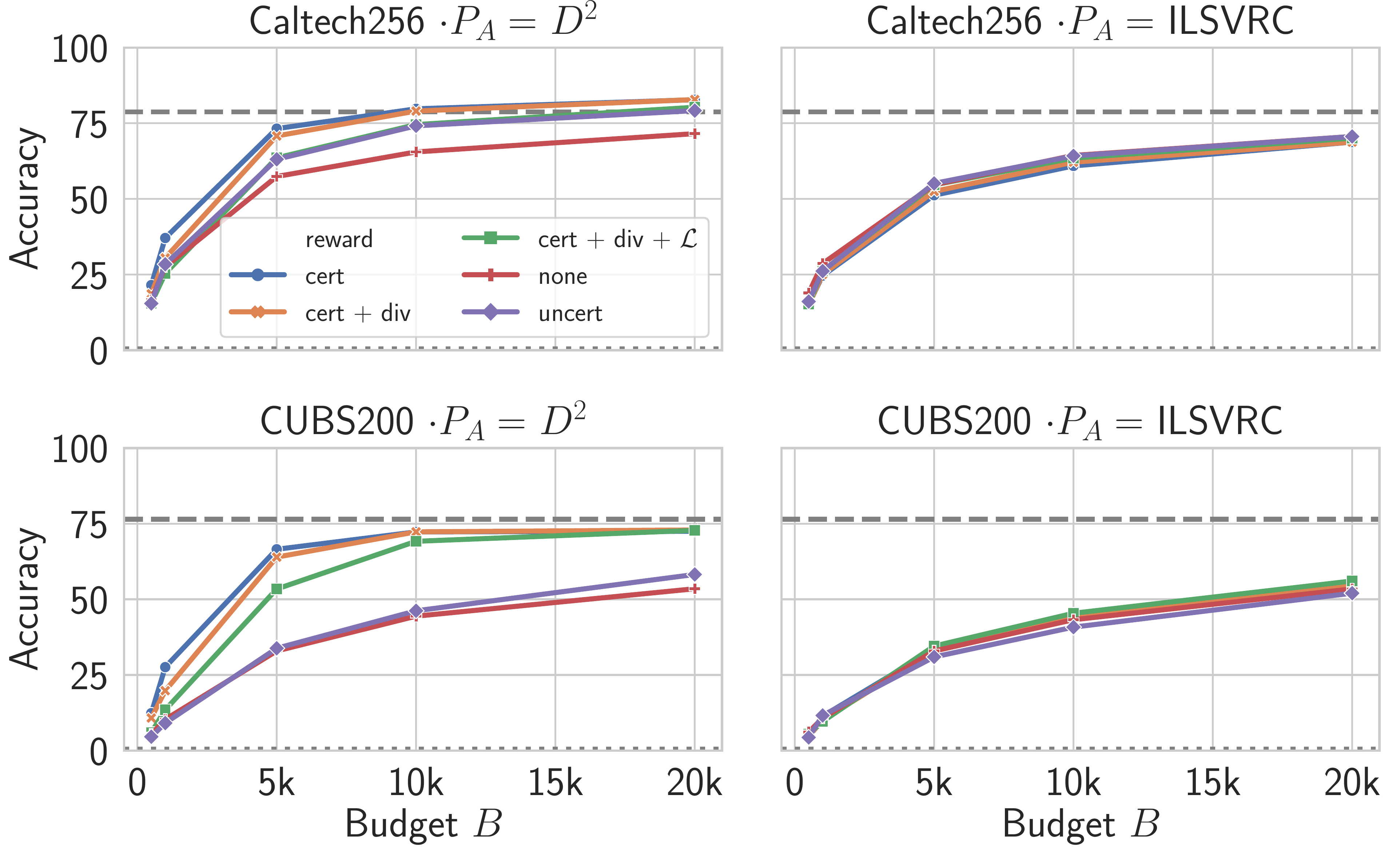}
    \end{center}
	\vspace{-1.5em}
   \caption{\textbf{Reward Ablation.} {\fontfamily{lmss}\selectfont cert}: certainty, {\fontfamily{lmss}\selectfont uncert}: uncertainty, {\fontfamily{lmss}\selectfont div}: diversity, $\loss$: loss, {\fontfamily{lmss}\selectfont none}: no reward (\texttt{random} strategy).}
	\label{fig:expt_reward_ablation}
	\vspace{-0.5em}
\end{figure}

\myparagraphnp{What makes for a good \textit{reward}?}
Using the \texttt{adaptive} sampling strategy, we now address influence of three rewards (discussed in Section \ref{sec:approach_query_adaptive}).
We observe:
(i) \textbf{closed-world} (Fig.~\ref{fig:expt_reward_ablation} left):
All reward signals in \texttt{adaptive} helps with the sample efficiency over \texttt{random}.
Reward \texttt{cert} (which encourages exploitation) provides the best feedback signal.
Including other rewards (\texttt{cert+div+$\loss$}) slightly deteriorates performance, as they encourage \textit{exploration} over related or unseen actions -- which is not ideal in a closed-world.
Reward \texttt{uncert}, a popular measure used in AL literature \cite{ebert2012ralf,beluch2018power,settles2008analysis} underperforms in our setting since it encourages uncertain (in our case, irrelevant) images.
(ii) \textbf{open-world} (Fig.~\ref{fig:expt_reward_ablation} right):
All rewards display only none-to-marginal improvements for all choices of $F_V$, with the highest improvement in \modelname{CUBS200} using \texttt{cert+div+$\loss$}.
However, we notice an influence on learnt policies where adopting exploration (\texttt{div + $\loss$}) with exploitation (\texttt{cert}) goals result in a softer probability distribution $\pi$ over the action space and in turn, encouraging related images.

\begin{figure}
    \begin{center}
       \includegraphics[width=\linewidth]{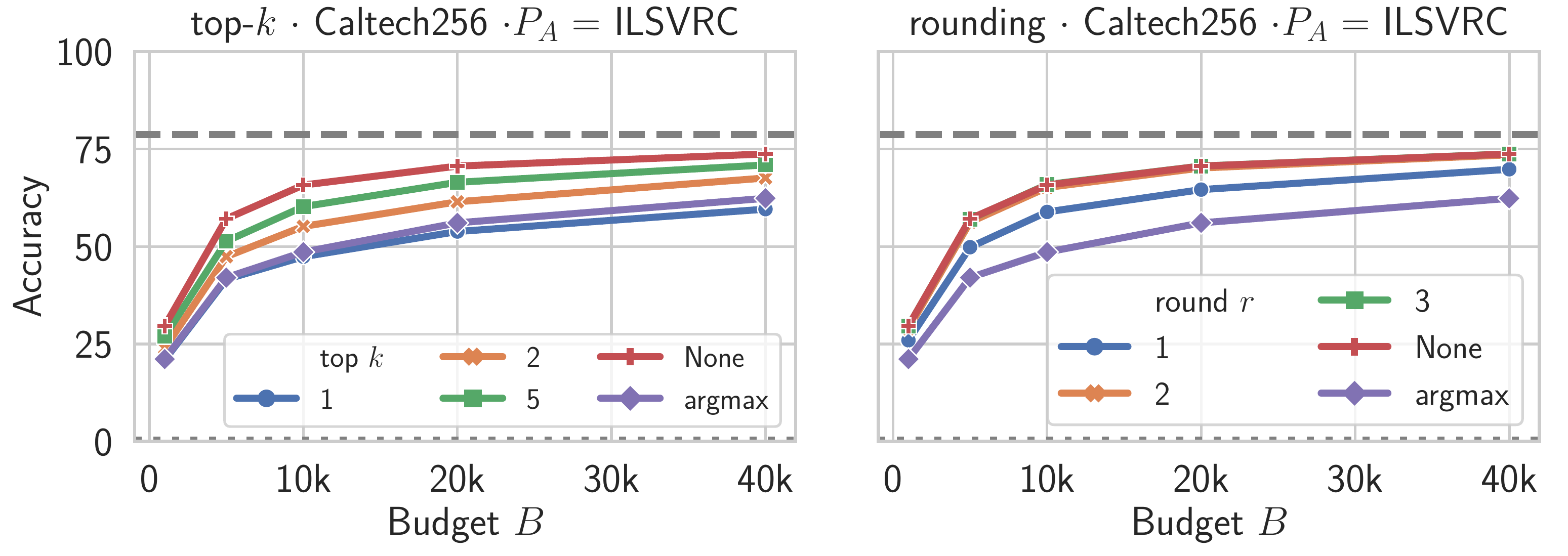}
    \end{center}
	\vspace{-1.0em}
   \caption{\textbf{Truncated Posteriors.} Influence of training knockoff with truncated posteriors.}
	\label{fig:expt_topk_rounding}
	\vspace{-0.5em}
\end{figure}

\myparagraphnp{Can we train knockoffs with \textit{truncated blackbox outputs}?}
So far, we found adversary's \textit{attack} objective of knocking off blackbox models can be effectively carried out with minimal assumptions.
Now we explore the influence of victim's \textit{defense} strategy of reducing informativeness of blackbox predictions to counter adversary's model stealing attack.
We consider two truncation strategies:
(a) top-$k$: top-$k$ (out of $K$) unnormalized posterior probabilities are retained, while rest are zeroed-out;
(b) rounding $r$: posteriors are rounded to $r$ decimals e.g., round(0.127, $r$=2) = 0.13.
In addition, we consider the extreme case ``argmax'', where only index $k = \argmax_k y_k$ is returned.

From Figure~\ref{fig:expt_topk_rounding} (with $K$ = 256), we observe:
(i) truncating $\vecy_i$ -- either using top-$k$ or rounding -- slightly impacts the knockoff performance, with argmax achieving 0.76-0.84$\times$ accuracy of original performance for any budget $B$;
(ii) top-$k$: even small increments of $k$ significantly recovers the original performance -- $0.91\times$ at $k=2$ and $0.96\times$ at $k=5$;
(iii) rounding: recovery is more pronounced, with $0.99\times$ original accuracy achieved at just $r=2$.
We find model functionality stealing minimally impacted by reducing informativeness of blackbox predictions.

\subsection{Architecture choice}
\begin{figure}
    \begin{center}
       \includegraphics[width=\linewidth]{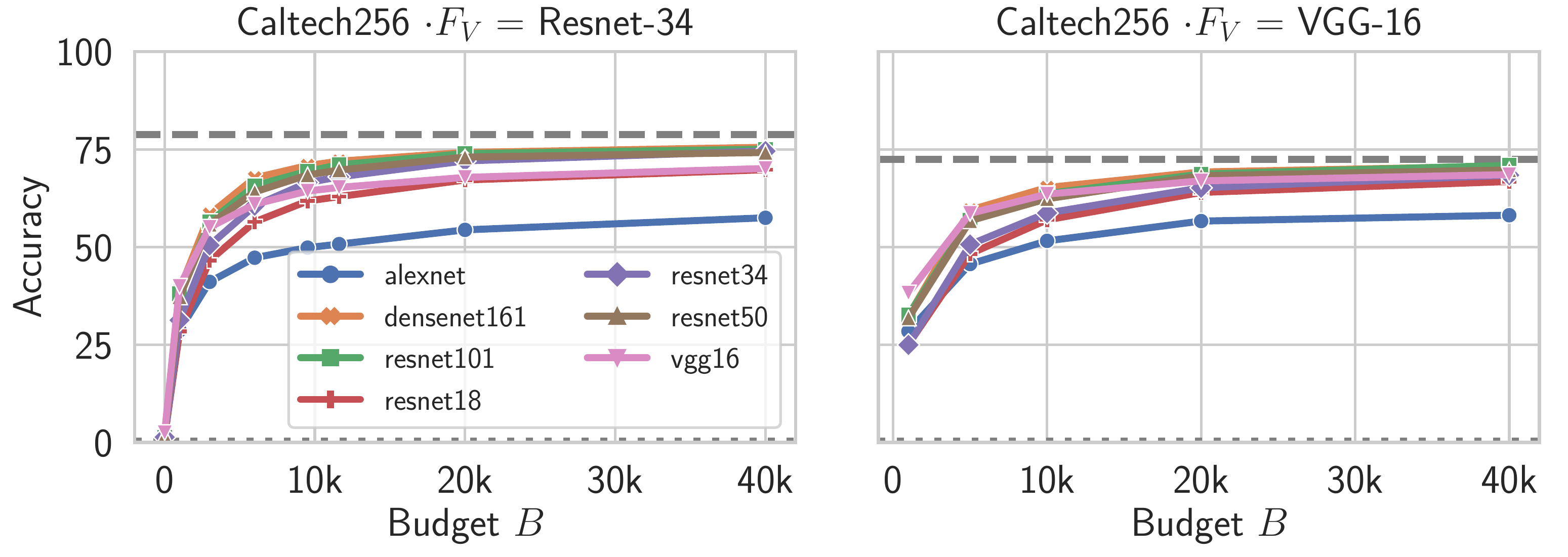}
    \end{center}
	\vspace{-1.0em}
   \caption{\textbf{Architecture choices.} $F_V$ (left: Resnet-34 and right: VGG-16) and $F_A$ (lines in each plot).}
	\label{fig:expt_fa}
	\vspace{-1.0em}
\end{figure}

In the previous section, we found model functionality stealing to be consistently effective while keeping the architectures of the blackbox and knockoff fixed.
Now we study the influence of the architectural choice $F_A$ vs. $F_V$.

\myparagraphnp{How does the \textit{architecture} of $F_A$ influence knockoff performance?}
We study the influence using two choices of the blackbox $F_V$ architecture: Resnet-34 \cite{he2016deep} and VGG-16 \cite{simonyan2014very}.
Keeping these fixed, we vary architecture of the knockoff $F_A$ by choosing from: Alexnet \cite{krizhevsky2012imagenet}, VGG-16 \cite{simonyan2014very}, Resnet-\{18, 34, 50, 101\} \cite{he2016deep}, and Densenet-161 \cite{huang2017densely}.

From Figure~\ref{fig:expt_fa}, we observe:
(i) performance of the knockoff ordered by model complexity: Alexnet (lowest performance) is at one end of the spectrum while significantly more complex Resnet-101/Densenet-161 are at the other;
(ii) performance transfers across model families: Resnet-34 achieves similar performance when stealing VGG-16 and vice versa;
(iii) complexity helps: selecting a more complex model architecture of the knockoff is beneficial. This contrasts KD settings where the objective is to have a more compact student (knockoff) model.

\subsection{Stealing Functionality of a Real-world Black-box Model}
Now we validate model functionality stealing on a popular image analysis API.
Such image recognition services are gaining popularity allowing users to obtain image-predictions for a variety of tasks at low costs (\$1-2 per 1k queries).
These image recognition APIs have also been used to evaluate other attacks e.g., adversarial examples \cite{liu2016delving,bhagoji2017exploring,ilyas2018black}.
We focus on a facial characteristics API which given an image, returns attributes and confidences per face.
Note that in this experiment, we have semantic information of blackbox output classes.

\myparagraph{Collecting $P_A$}
The API returns probability vectors per face in the image and thus, querying irrelevant images leads to a wasted result with no output information.
Hence, we use two face image sets $P_A$ for this experiment: CelebA (220k images) \cite{liu2015faceattributes} and OpenImages-Faces (98k images).
We create the latter by cropping faces (plus margin) from images in the OpenImages dataset  \cite{kuznetsova2018open}.

\myparagraph{Evaluation}
Unlike previous experiments, we cannot access victim's test data.
Hence, we create test sets for each image set by collecting and manually screening seed annotations from the API on $\sim$5K images.

\begin{figure}
    \begin{center}
       \includegraphics[width=\linewidth]{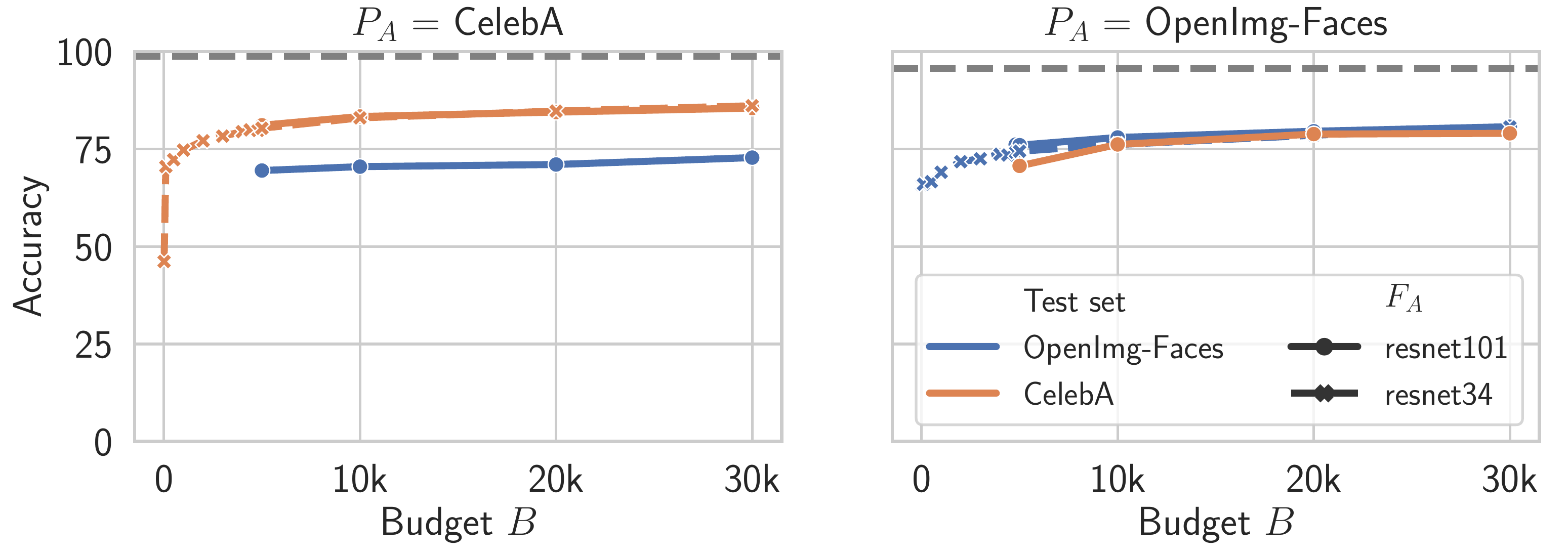}
    \end{center}
	\vspace{-1.0em}
   \caption{\textbf{Knocking-off a real-world API.} Performance of the knockoff achieved with two choices of $P_A$.}
	\label{fig:expt_real_api}
	\vspace{-1.0em}
\end{figure}

\myparagraphnp{How does this translate to the \textit{real-world}?}
We model two variants of the knockoff using the \texttt{random} strategy (\texttt{adaptive} is not used since no relevant auxiliary information of images are available).
We present each variant using two choices of architecture $F_A$: a compact Resnet-34 and a complex Resnet-101.
From Figure~\ref{fig:expt_real_api}, we observe:
(i) strong performance of the knockoffs achieving $0.76$-$0.82\times$ performance as that of the API on the test sets;
(ii) the diverse nature OpenImages-Faces helps improve generalization resulting in $0.82\times$ accuracy of the API on both test-sets;
(iii) the complexity of $F_A$ does not play a significant role: both Resnet-34 and Resnet-101 show similar performance indicating a compact architecture is sufficient to capture discriminative features for this particular task.

We find model functionality stealing translates well to the real-world with knockoffs exhibiting a strong performance.
The knockoff circumvents  monetary and labour costs of:
(a) collecting images for the task;
(b) obtaining expert annotations; and
(c) tuning a model.
As a result, an inexpensive knockoff is trained which exhibits strong performance, using victim API queries amounting to only \$30.

\section{Conclusion}
We investigated the problem of model functionality stealing where an adversary transfers the functionality of a victim model into a knockoff via blackbox access.
In spite of minimal assumptions on the blackbox, we demonstrated the surprising effectiveness of our approach.
Finally, we validated our approach on a popular image recognition API and found strong performance of knockoffs.
We find functionality stealing poses a real-world threat that potentially undercuts an increasing number of deployed ML models.

\myparagraph{Acknowledgement}
This research was partially supported by the German Research Foundation (DFG CRC 1223).
We thank Yang Zhang for helpful discussions.


{\small
\bibliographystyle{ieee}
\bibliography{main}
}


\clearpage
\appendix
\noindent \textbf{\LARGE Appendices}

\section{Contents}
The appendix contains:
\begin{enumerate}[label=\Alph*.]
    \item Contents (this section)
    
    \item Extended descriptions
    \begin{enumerate}[label=\arabic*.]
     \item Blackbox models
     \item Overlap between $P_V$ and $P_A$
     \item Aggregating OpenImages and OpenImages-Faces
     \item Additional implementation details
   \end{enumerate}
   
    \item Extensions of existing results
    \begin{enumerate}[label=\arabic*.]
     \item Qualitative Results
     \item Sample-efficiency of GT
     \item Policies learnt by \texttt{adaptive} strategy
     \item Reward Ablation
   \end{enumerate}
   
    \item Auxiliary experiments
    \begin{enumerate}[label=\arabic*.]
     \item Seen and unseen classes
     \item \texttt{Adaptive} strategy: With/without hierarchy
     \item Semi-open world: $\tau D^2$
   \end{enumerate}
\end{enumerate}

\section{Extended Descriptions}
\label{sec:app_ext}

In this section, we provide additional detailed descriptions and implementation details.

\subsection{Black-box models}
\label{sec:app_ext_bbox}

We supplement Section \ref{sec:exptsetup_fv} by providing extended descriptions of the blackboxes listed in Table \ref{tab:blackboxes}.
Each blackbox $F_V$ is trained on one particular image classification dataset.

\myparagraph{Black-box 1: \modelname{Caltech256} \cite{griffin2007caltech}}
Caltech-256 is a popular dataset for general object recognition gathered by downloading relevant examples from Google Images and manually screening for quality and errors.
The dataset contains 30k images covering 256 common object categories.

\myparagraph{Black-box 2: \modelname{CUBS200} \cite{WahCUB_200_2011}}
A fine-grained bird-classifier is trained on the CUBS-200-2011 dataset.
This dataset contains roughly 30 train and 30 test images for each of 200 species of birds.
Due to the low intra-class variance, collecting and annotating images is challenging even for expert bird-watchers.

\myparagraph{Black-box 3: \modelname{Indoor67} \cite{quattoni2009recognizing}}
We introduce another fine-grained task of recognizing 67 types of indoor scenes.
This dataset consists of 15.6k images collected from Google Images, Flickr, and LabelMe.

\myparagraph{Black-box 4: \modelname{Diabetic5} \cite{eyepacs}}
Diabetic Retinopathy (DR) is a medical eye condition characterized by retinal damage due to diabetes.
Cases are typically determined by trained clinicians who look for presence of lesions and vascular abnormalities in digital color photographs of the retina captured using specialized cameras.
Recently, a dataset of such 35k retinal image scans was made available as a part of a Kaggle competition \cite{eyepacs}.
Each image is annotated by a clinician on a scale of 0 (no DR) to 4 (proliferative DR).
This highly-specialized biomedical dataset also presents challenges in the form of extreme imbalance (largest class contains 30$\times$ as the smallest one).

\subsection{Overlap: Open-world}
\label{sec:app_ext_overlap}

\begin{table}[]
\centering
\scriptsize
\begin{tabular}{@{}lllll@{}}
\toprule
          & \multicolumn{4}{c}{$P_V$} \\ \cmidrule{2-5}
$P_A$           & \begin{tabular}[c]{@{}c@{}}\modelname{Caltech256}\\ ($K$=256)\end{tabular} & \begin{tabular}[c]{@{}c@{}}\modelname{CUBS200}\\ ($K$=200)\end{tabular} & \begin{tabular}[c]{@{}c@{}}\modelname{Indoor67}\\ ($K$=67)\end{tabular} & \begin{tabular}[c]{@{}c@{}}\modelname{Diabetic5}\\ ($K$=5)\end{tabular} \\ \midrule
ILSVRC ($Z$=1000)     & 108 (42\%)        & 2 (1\%)    & 10 (15\%)      & 0 (0\%)       \\
OpenImages ($Z$=601) & 114 (44\%)        & 1 (0.5\%)    & 4 (6\%)      & 0 (0\%)       \\
\bottomrule
\end{tabular}
\caption{\textbf{Overlap between $P_A$ and $P_V$.}}
\label{tab:overlap}
\end{table}

\begin{figure*}[h]
    \begin{center}
       \includegraphics[width=\linewidth]{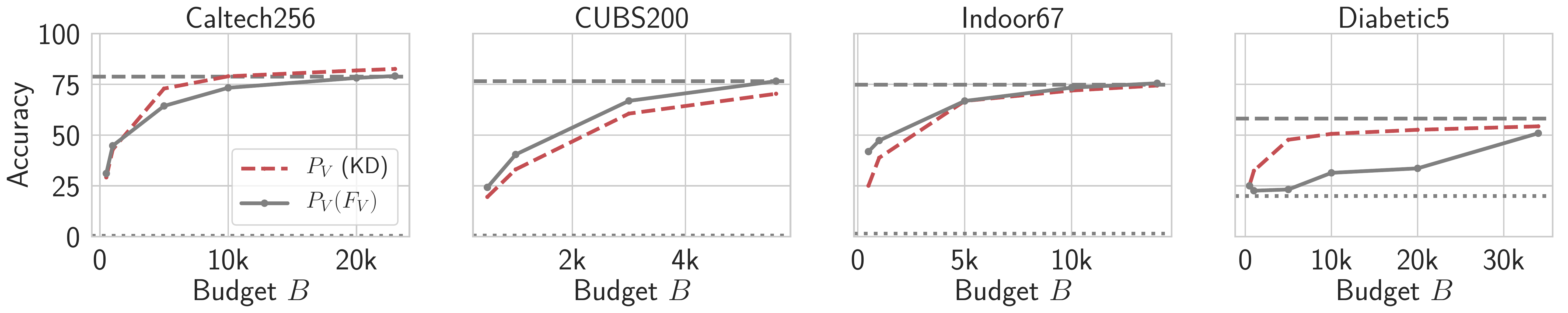}
    \end{center}
   \caption{\textbf{Training on GT vs. KD.} Extension of Figure 5. We compare sample efficiency of first two rows in Table \ref{tab:pa_summary}: ``$P_V (F_V)$'' (training with GT data) and ``$P_V$ (KD)'' (training with soft-labels of GT images produced by $F_V$)}
	\label{fig:expt_kd_gt}
\end{figure*}

In this section, we supplement Section \ref{sec:exptsetup_pa_choice} in the main paper by providing more details on how overlap was calculated in the open-world scenarios.
We manually compute overlap between labels of the blackbox ($K$, e.g., 256 Caltech classes) and the adversary's dataset ($Z$, e.g., 1k ILSVRC classes) as: $100 \times |K \cap Z| / |K|$.
We denote two labels $k \in K$ and $z \in Z$ to overlap if:
(a) they have the same semantic meaning; or
(b) $z$ is a type of $k$ e.g., $z$ = ``maltese dog'' and $k$ = ``dog''.
The exact numbers are provided in Table \ref{tab:overlap}.
We remark that this is a soft-lower bound.
For instance, while ILSVRC contains ``Hummingbird'' and CUBS-200-2011 contains three distinct species of hummingbirds, this is not counted towards the overlap as the adversary lacks annotated data necessary to discriminate among the three species.

\subsection{Dataset Aggregation}
\label{sec:app_ext_dataagg}

All datasets used in the paper (expect OpenImages) have been used in the form made publicly available by the authors.
We use a subset of OpenImages due to storage constraints imposed by its massive size (9M images).
The description to obtain these subsets are provided below.

\myparagraph{OpenImages}
We retrieve 2k images for each of the 600 OpenImages \cite{kuznetsova2018open} ``boxable'' categories, resulting in 554k unique images.
$\sim$19k images are removed for either being corrupt or representing Flickr's placeholder for unavailable images.
This results in a total of 535k unique images.

\myparagraph{OpenImages-Faces}
We download all images (422k) from OpenImages \cite{kuznetsova2018open} with label ``\texttt{/m/0dzct: Human face}'' using the OID tool \cite{OIDv4_ToolKit}.
The bounding box annotations are used to crop faces (plus a margin of 25\%) containing at least 180$\times$180 pixels.
We restrict to at most 5 faces per image to maintain diversity between train/test splits.
This results in a total of 98k faces images.

\subsection{Additional Implementation Details}
\label{sec:app_ext_impl}

In this section, we provide implementation details to supplement discussions in the main paper.

\myparagraph{Training $F_V = $ \modelname{Diabetic5}}
Training this victim model is identical to other blackboxes except for one aspect: weighted loss.
Due to the extreme imbalance between classes of the dataset, we weigh each class as follows.
Let $n_k$ denote the number of images belonging to class $k$ and let $n_{\min} = \min_k n_k$.
We weigh the loss for each class $k$ as $n_{\min} / n_k$.
From our experiments with weighted loss, we found approximately 8\% absolute improvement in overall accuracy on the test set.
However, the training of knockoffs of all blackboxes are identical in all aspects, including a non-weighted loss irrespective of the victim blackbox targeted.

\myparagraph{Creating ILSVRC Hierarchy}
We represent the 1k labels of ILSVRC as a hierarchy Figure \ref{fig:agent}b in the form: root node ``entity'' $\rightarrow$ $N$ coarse nodes $\rightarrow$ 1k leaf nodes.
We obtain $N$ (30 in our case) coarse labels as follows:
(i) a 2048-d mean feature vector representation per 1k labels is obtained using an Imagenet-pretrained ResNet ;
(ii) we cluster the 1k features into $N$ clusters using scikit-learn's \cite{scikit-learn} implementation of agglomerative clustering;
(iii) we obtain semantic labels per cluster (i.e., coarse node) by finding the common parent in the Imagenet semantic hierarchy.

\myparagraph{\texttt{Adaptive} Strategy}
Recall from Section \ref{sec:results}, we train the knockoff in two phases:
(a) \textit{Online}: during transfer set construction; followed by
(b) \textit{Offline}: the model is retrained using transfer set obtained thus far.
In phase (a), we train $F_A$ with SGD (with 0.5 momentum) with a learning rate of 0.0005 and batch size of 4 (i.e., 4 images sampled at each $t$).
In phase (b), we train the knockoff $F_A$ from scratch on the transfer set using SGD (with 0.5 momentum) for 100 epochs with learning rate of 0.01 decayed by a factor of 0.1 every 60 epochs. 

\input{qual_results.tex}

\begin{figure*}[t]
    \centering
    \begin{subfigure}[b]{1.0\linewidth}
         \centering
         \includegraphics[width=\linewidth]{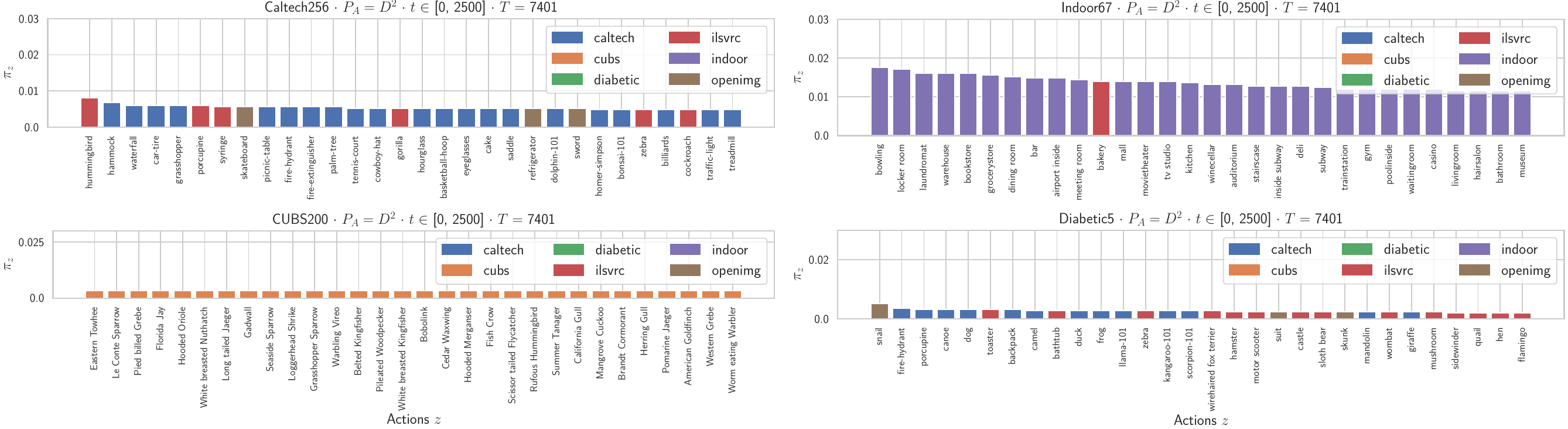}
         \caption{\textbf{Closed world.}}
         \vspace{3.0em}
         \label{fig:expt_policy_d2_static}
     \end{subfigure}
     \hfill
    \begin{subfigure}[b]{1.0\linewidth}
         \centering
         \includegraphics[width=\linewidth]{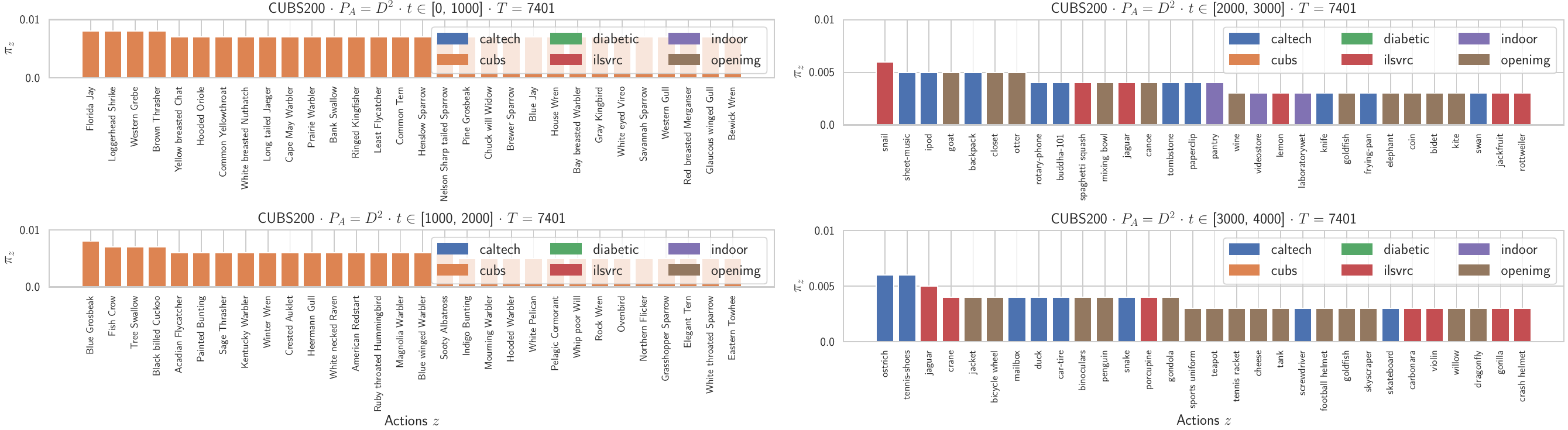}
         \caption{\textbf{Closed world.} Analyzing policy over time $t$ for \modelname{CUBS200}.}
         \vspace{3.0em}
         \label{fig:expt_policy_d2_dynamic}
     \end{subfigure}
     \hfill
    \begin{subfigure}[b]{1.0\linewidth}
         \centering
         \includegraphics[width=\linewidth]{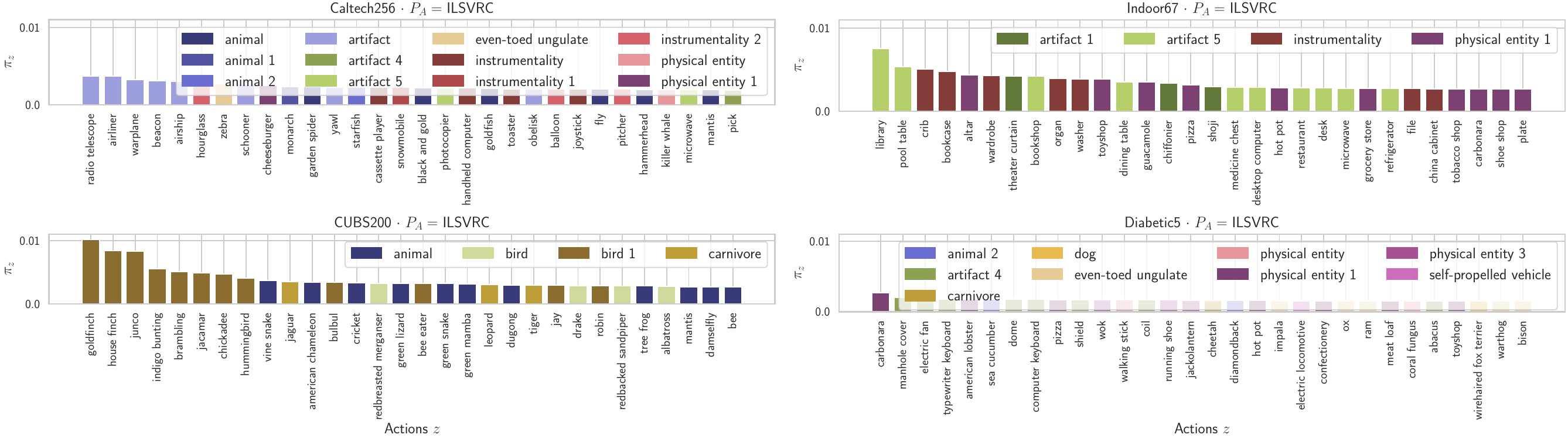}
         \caption{\textbf{Open world.}}
         \vspace{1.0em}
         \label{fig:expt_policy_ilsvrc_static}
     \end{subfigure}
   \caption{\textbf{Policies learnt by \texttt{adaptive} strategy.} Supplements Figure \ref{fig:expt_policy} in the main paper.}
	\label{fig:expt_policies}
	\vspace{-1.0em}
\end{figure*}

\begin{figure}[t]
    \begin{center}
       \includegraphics[width=\linewidth]{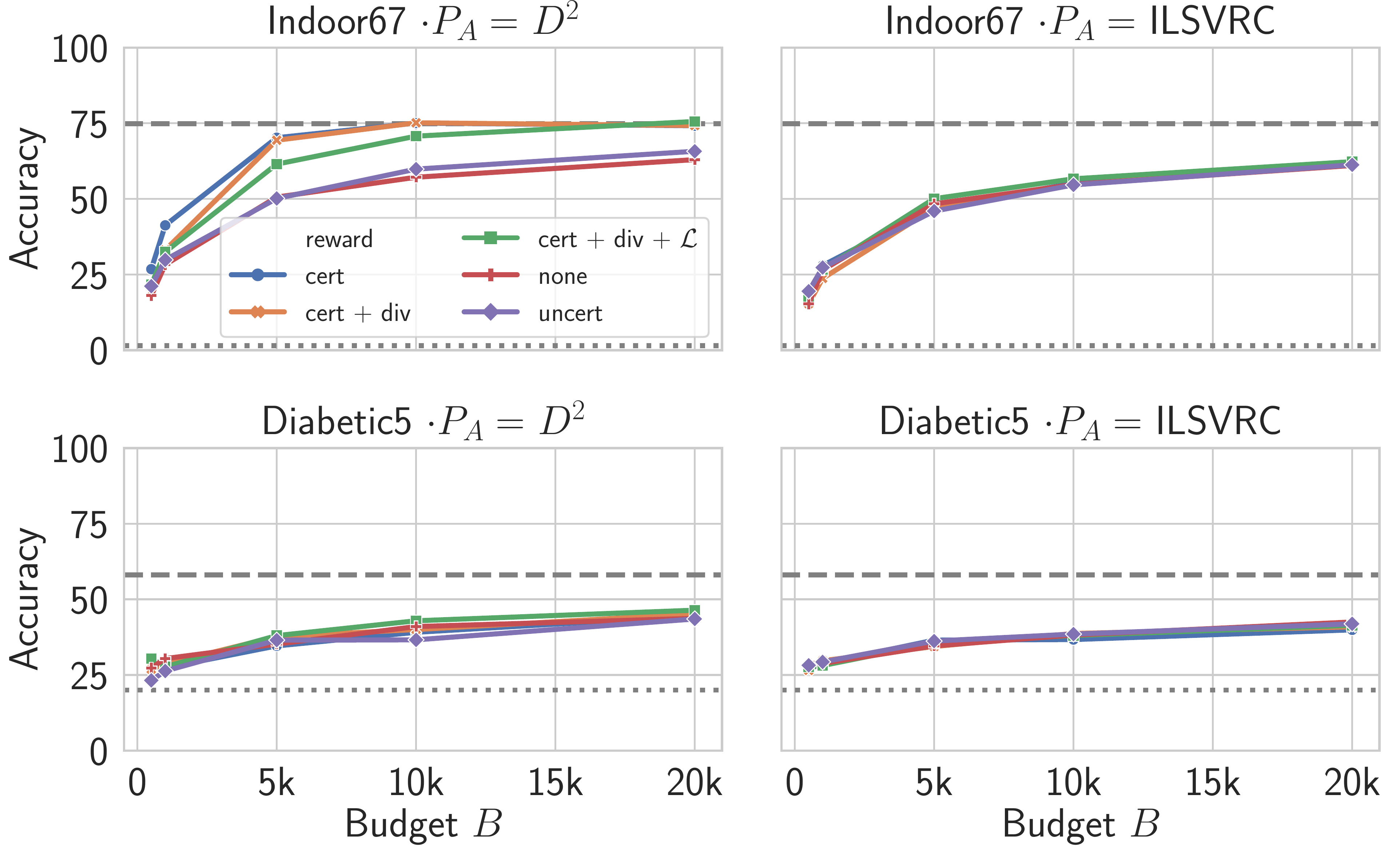}
    \end{center}
	\vspace{-1.5em}
   \caption{\textbf{Reward Ablation.} Supplements Figure \ref{fig:expt_reward_ablation} in the main paper.}
	\label{fig:expt_reward_ablation_2}
	\vspace{-1.0em}
\end{figure}

\begin{figure}[t]
    \begin{center}
       \includegraphics[width=\linewidth]{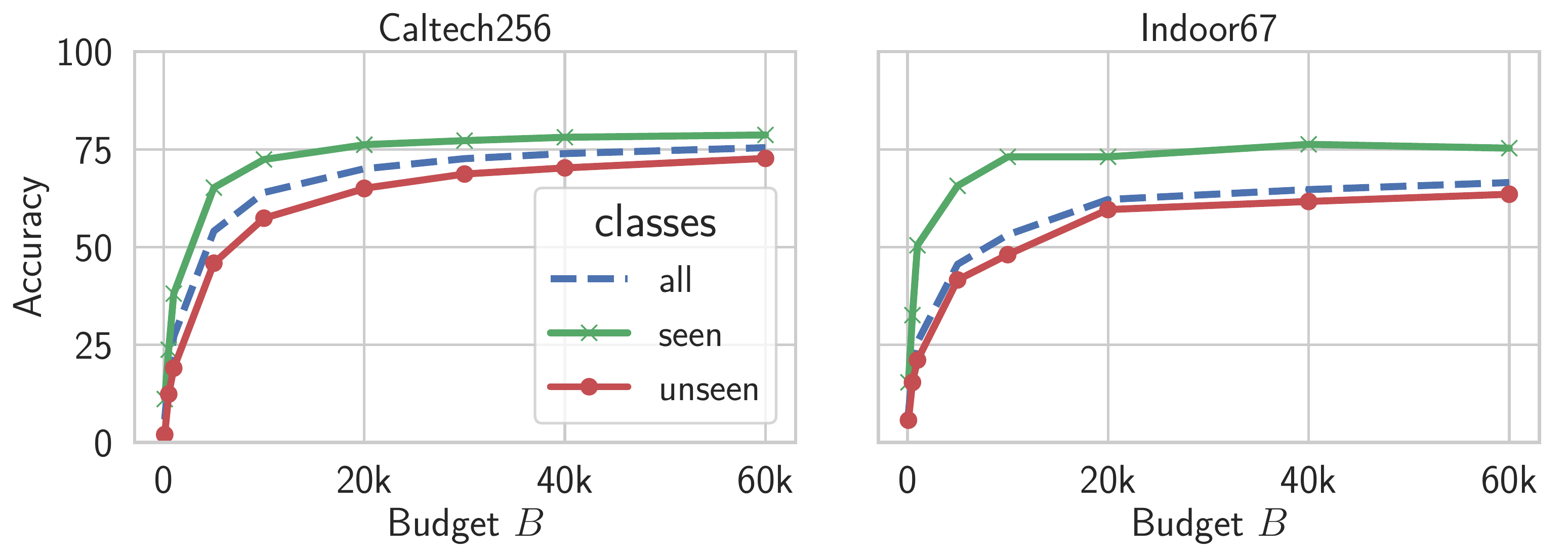}
    \end{center}
	\vspace{-1.5em}
   \caption{\textbf{Per class evaluation.} Per-class evaluation split into seen and unseen classes.}
	\label{fig:expt_seen_unseen}
	\vspace{-1.0em}
\end{figure}

\section{Extensions of Existing Results}
\label{sec:app_results}
In this section, we present extensions of existing results discussed in the main paper.

\subsection{Qualitative Results}
\label{sec:app_results_qual}

Qualitative results to supplement Figure \ref{fig:expt_qual} are provided in Figures \ref{tab:expt_qual_caltech}-\ref{tab:expt_qual_diabetic}.
Each row in the figures correspond to an output class of the blackbox whose images the knockoff has never encountered before.
Images in the ``transfer set'' column were randomly sampled from ILSVRC \cite{russakovsky2015imagenet,deng2009imagenet}.
In contrast, images in the ``test set'' belong to the victim's test set (Caltech256, CUBS-200-2011, etc.).

\subsection{Sample Efficiency: Training Knockoffs on GT}
\label{sec:app_results_kdgt}

We extend Figure \ref{fig:expt_pa} in the main paper to include training on the same ground-truth data used to train the blackboxes.
This extension ``$P_V (F_V)$'' is illustrated in Figure \ref{fig:expt_kd_gt}, displayed alongside KD approach.
The figure represents the sample-efficiency of the first two rows of \ref{tab:pa_summary}.
Here we observe:
(i) comparable performance in all but one case (\modelname{Diabetic5}, discussed shortly) indicating KD is an effective approach to train knockoffs;
(ii) we find KD achieve better performance in \modelname{Caltech256} and \modelname{Diabetic5} due to regularizing effect of training on soft-labels \cite{hinton2015distilling} on an imbalanced dataset.

\subsection{Policies learnt by \texttt{Adaptive}}
\label{sec:app_results_policies}

We inspected the policy $\pi$ learnt by the \texttt{adaptive} strategy in Section 6.1.
In this section, we provide policies over all blackboxes in the closed- and open-world setting.
Figures \ref{fig:expt_policy_d2_static} and \ref{fig:expt_policy_ilsvrc_static} display probabilities of each action $z \in Z$ at $t=2500$.

Since the distribution of rewards is non-stationary, we visualize the policy over time in Figure \ref{fig:expt_policy_d2_dynamic} for \modelname{CUBS200} in a closed-world setup.
From this figure, we observe an evolution where:
(i) at early stages ($t \in [0, 2000]$), the approach samples (without replacement) images that overlaps with the victim's train data; and
(ii) at later stages ($t \in [2000, 4000]$), since the overlapping images have been exhausted, the approach explores related images from other datasets e.g., ``ostrich'', ``jaguar''.

\subsection{Reward Ablation}
\label{sec:app_results_reward}

The reward ablation experiment \ref{fig:expt_reward_ablation} for the remaining datasets are provided in Figure \ref{fig:expt_reward_ablation_2}.
We make similar observations as before for \modelname{Indoor67}.
However, since $F_V = $ \modelname{Diabetic5} demonstrates confident predictions in all images (discussed in 563-567), we find little-to-no improvement for knockoffs of this victim model.


\section{Auxiliary Experiments}
\label{sec:app_expts}

In this section, we present experiments to supplement existing results in the main paper.

\begin{figure*}[t]
    \begin{center}
       \includegraphics[width=\linewidth]{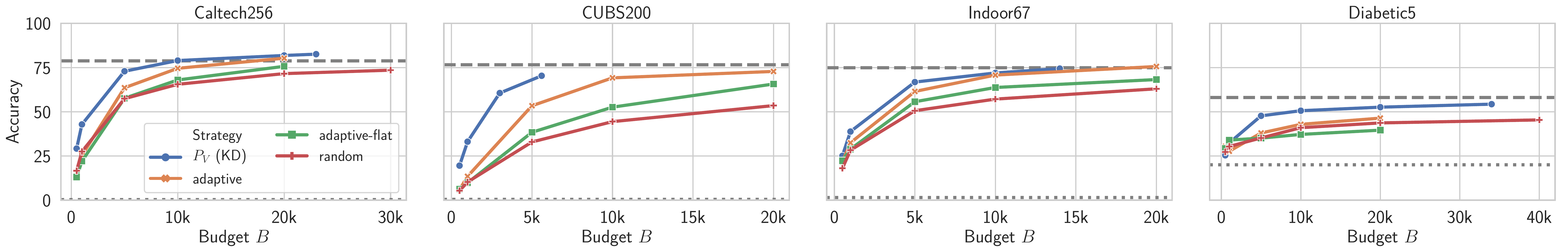}
    \end{center}
   \caption{\textbf{Hierarchy.} Evaluating \texttt{adaptive} with and without hierarchy using $P_A = D^2$. 
   {\color{gray}{\hdashrule[0.5ex]{0.4cm}{1pt}{1mm 1.2pt}}} represents accuracy of blackbox $F_V$ and {\color{gray}{\hdashrule[0.5ex]{0.4cm}{1pt}{1pt}}} represents chance-level performance.}
	\label{fig:expt_hierarchy}
\end{figure*}

\begin{figure*}[t]
    \begin{center}
       \includegraphics[width=\linewidth]{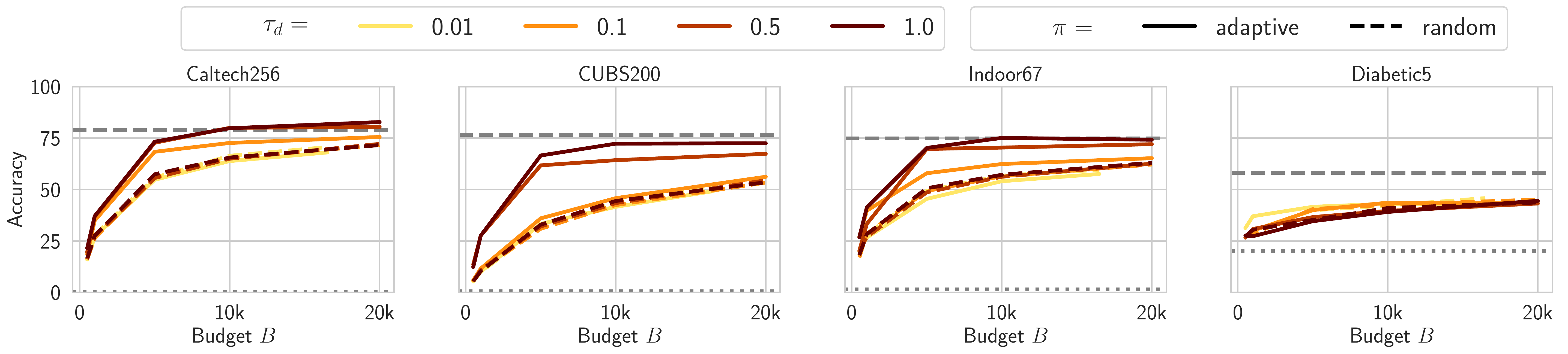}
       \includegraphics[width=\linewidth]{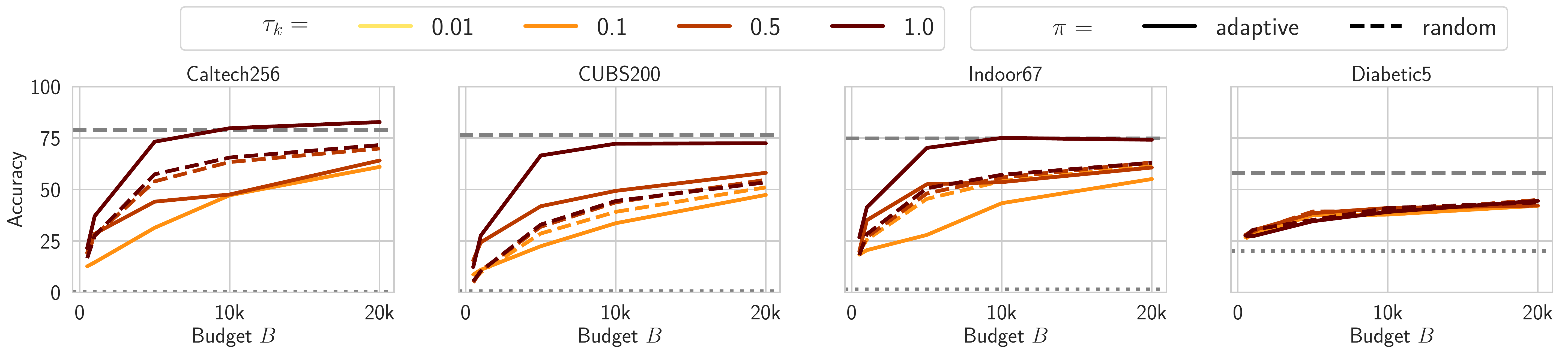}
    \end{center}
   \caption{\textbf{Semi-open world: $\tau_d$ and $\tau_k$.}}
	\label{fig:expt_tau_d}
\end{figure*}


\subsection{Seen and Unseen classes}
\label{sec:app_expts_seen}

We now discuss evaluation to supplement Section \ref{sec:exptsetup_pa_choice} and Section \ref{sec:results}.

In \ref{sec:results_transfer} we highlighted strong performance of the knockoff even among classes that were never encountered (see Table \ref{tab:overlap} for exact numbers) during training.
To elaborate, we split the blackbox output classes into ``seen'' and ``unseen'' categories and present mean per-class accuracies in Figure \ref{fig:expt_seen_unseen}.
Although we find better performance on classes seen while training the knockoff, performance of unseen classes is remarkably high, with the knockoff achieving $>$70\% performance in both cases.

\subsection{\texttt{Adaptive}: With and without hierarchy}
\label{sec:app_expts_hierarchy}

The \texttt{adaptive} strategy presented in Section \ref{sec:approach_query_adaptive} uses a hierarchy discussed in Section 5.2.2.
As a result, we approached this as a hierarchical multi-armed bandit problem.
Now, we present an alternate approach \texttt{adaptive-flat}, without the hierarchy.
This is simply a multi-armed bandit problem with $|Z|$ arms (actions).

Figure \ref{fig:expt_hierarchy} illustrates the performance of these approaches using $P_A = D^2$ ($|Z|$ = 2129) and rewards \{certainty, diversity, loss\}.
We observe \texttt{adaptive} consistently outperforms \texttt{adaptive-flat}.
For instance, in \modelname{CUBS200}, \texttt{adaptive} is 2$\times$ more sample-efficient to reach accuracy of 50\%.
We find the hierarchy helps the adversary (agent) better navigate the large action space.

\subsection{Semi-open World}
\label{sec:app_expts_semiopen}
The closed-world experiments ($P_A = D^2$) presented in Section \ref{sec:results_transfer} and discussed in Section \ref{sec:exptsetup_pa_choice} assumed access to the image universe.
Thereby, the overlap between $P_A$ and $P_V$ was 100\%.
Now, we present an intermediate overlap scenario \textbf{semi-open world} by parameterizing the overlap as:
(i) $\tau_\text{d}$: The overlap between \textit{images} $P_A$ and $P_V$ is $100 \times \tau_\text{d}$; and
(ii) $\tau_\text{k}$: The overlap between \textit{labels} $K$ and $Z$ is $100 \times \tau_\text{k}$.
In both these cases $\tau_\text{d}, \tau_\text{k} \in (0, 1]$ represents the fraction of $P_A$ used.
$\tau_\text{d} = \tau_\text{k} = 1$ depicts the closed-world scenario discussed in Section 6.1.

From Figure \ref{fig:expt_tau_d}, we observe:
(i) the \texttt{random} strategy is unaffected in the semi-open world scenario, displaying comparable performance for all values of $\tau_\text{d}$ and $\tau_\text{k}$;
(ii) $\tau_\text{d}$: knockoff obtained using \texttt{adaptive} obtains strong performance even with low overlap e.g., a difference of at most 3\% performance in \modelname{Caltech256} even at $\tau_\text{d} = 0.1$;
(iii) $\tau_\text{k}$: although the \texttt{adaptive} strategy is minimally affected in few cases (e.g., \modelname{CUBS200}), we find the performance drop due to a pure exploitation (certainty) that is used. We observed recovery in performance by using all rewards indicating exploration goals (diversity, loss) are necessary when transitioning to an open-world scenario.

\end{document}

%% file: qual_results.tex
\begin{figure*}[]
\footnotesize
\centering
\begin{tabular}{ccccccc}
\toprule
\multicolumn{3}{c}{{\normalsize Transfer Set}} &  & \multicolumn{3}{c}{{\normalsize Test Set}} \\ \cmidrule{1-3} \cmidrule{5-7}
\includegraphics[width=2.3cm]{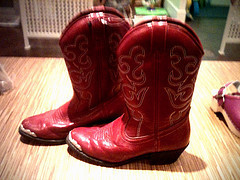} &
\includegraphics[width=2.3cm]{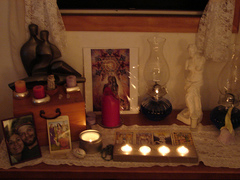} &         
\includegraphics[width=2.3cm]{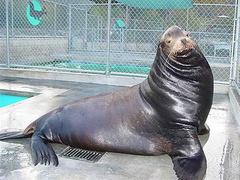} &  &      
\includegraphics[width=2.3cm]{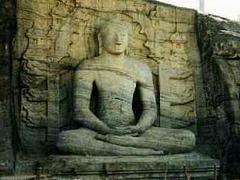} &        
\includegraphics[width=2.3cm]{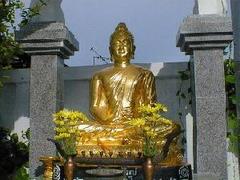} &
\includegraphics[width=2.3cm]{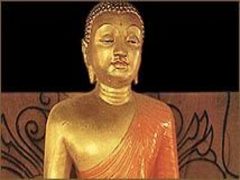} \\
\begin{tabular}[c]{@{}c@{}}Buddha: 0.65\\ Boxing glove: 0.07\\ Jesus Christ: 0.034 \\ $\;$ \end{tabular} &
\begin{tabular}[c]{@{}c@{}}Buddha: 0.49\\ Cake: 0.06\\ Minotaur: 0.05 \\ $\;$ \end{tabular} &
\begin{tabular}[c]{@{}c@{}}Buddha: 0.28\\ Minotaur: 0.09\\ Dog: 0.09 \\ $\;$ \end{tabular} & &
\begin{tabular}[c]{@{}c@{}}\underline{Buddha}: \hlg{0.68}\\ Tombstone: 0.07\\ Elephant: 0.04 \\ $\;$ \end{tabular} &
\begin{tabular}[c]{@{}c@{}}\underline{Buddha}: \hlg{0.5}\\ Minaret: 0.08\\ Tower Pisa: 0.05 \\ $\;$ \end{tabular} &
\begin{tabular}[c]{@{}c@{}}People: \hlr{0.26}\\ \underline{Buddha}: 0.25\\ T-shirt: 0.14 \\ $\;$ \end{tabular} \\

\includegraphics[width=2.3cm]{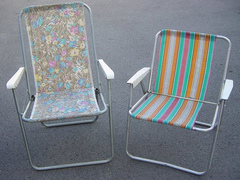} &
\includegraphics[width=2.3cm]{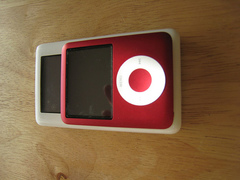} &         
\includegraphics[width=2.3cm]{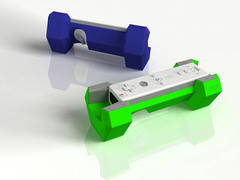} &  &      
\includegraphics[width=2.3cm]{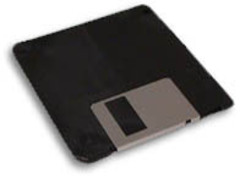} &        
\includegraphics[width=2.3cm]{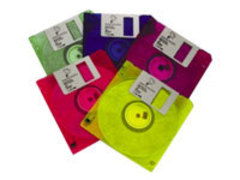} &
\includegraphics[width=2.3cm]{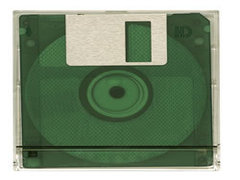} \\
\begin{tabular}[c]{@{}c@{}}Floppy-disk: 0.77\\ Socks: 0.04\\ Mattress: 0.03 \\ $\;$ \end{tabular} &
\begin{tabular}[c]{@{}c@{}}Floppy-disk: 0.46\\ iPod: 0.18\\ CD: 0.06 \\ $\;$ \end{tabular} &
\begin{tabular}[c]{@{}c@{}}Floppy-disk: 0.18\\ Flashlight: 0.08\\ Pez-dispenser: 0.07 \\ $\;$ \end{tabular} & &
\begin{tabular}[c]{@{}c@{}}\underline{Floppy-disk}: \hlg{0.74}\\ Necktie: 0.04\\ Video Projec.: 0.02 \\ $\;$ \end{tabular} &
\begin{tabular}[c]{@{}c@{}}\underline{Floppy-disk}: \hlg{0.61}\\ Socks: 0.02\\ Teddy bear: 0.02 \\ $\;$ \end{tabular} &
\begin{tabular}[c]{@{}c@{}}iPod: \hlr{0.49}\\ \underline{Floppy-disk}: 0.08\\ CD: 0.04 \\ $\;$ \end{tabular} \\

\includegraphics[width=2.3cm]{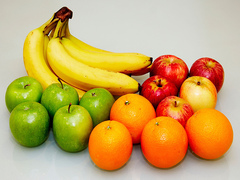} &
\includegraphics[width=2.3cm]{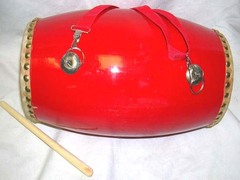} &         
\includegraphics[width=2.3cm]{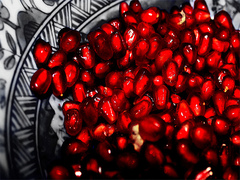} &  &      
\includegraphics[width=2.3cm]{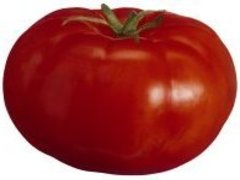} &        
\includegraphics[width=2.3cm]{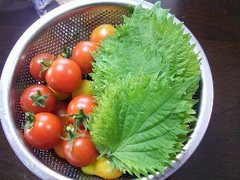} &
\includegraphics[width=2.3cm]{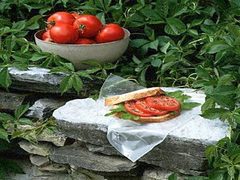} \\
\begin{tabular}[c]{@{}c@{}}Tomato: 0.96\\ Grapes: 0.01\\ Boxing glove: 0.00 \end{tabular} &
\begin{tabular}[c]{@{}c@{}}Tomato: 0.43\\ Welding mask: 0.12\\ Bowling ball: 0.01 \end{tabular} &
\begin{tabular}[c]{@{}c@{}}Tomato: 0.27\\ Dice: 0.18\\ Stained glass: 0.01 \end{tabular} & &
\begin{tabular}[c]{@{}c@{}}\underline{Tomato}: \hlg{0.94}\\ Boxing glove: 0.04\\ Bowling ball: 0.00 \end{tabular} &
\begin{tabular}[c]{@{}c@{}}\underline{Tomato}: \hlg{0.41}\\ Cactus: 0.07\\ Iguana: 0.6 \end{tabular} &
\begin{tabular}[c]{@{}c@{}}Mushroom: \hlr{0.17}\\ \underline{Tomato}: 0.10\\ Birdbath: 0.07 \end{tabular} \\
\bottomrule
\end{tabular}
\caption{\textbf{Qualitative results: \modelname{Caltech256}.} Extends Figure 6 in the main paper. \underline{GT} labels are underlined, \hlg{correct} knockoff top-1 predictions in green and \hlr{incorrect} in red.}
\label{tab:expt_qual_caltech}
\end{figure*}

\begin{figure*}[]
\footnotesize
\centering
\begin{tabular}{ccccccc}
\toprule
\multicolumn{3}{c}{{\normalsize Transfer Set}} &  & \multicolumn{3}{c}{{\normalsize Test Set}} \\ \cmidrule{1-3} \cmidrule{5-7}
\includegraphics[width=2.3cm]{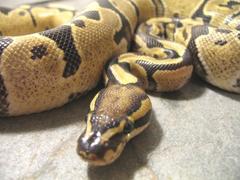} &
\includegraphics[width=2.3cm]{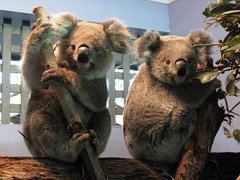} &         
\includegraphics[width=2.3cm]{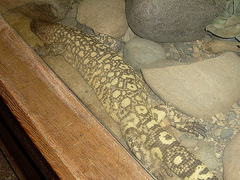} &  &      
\includegraphics[width=2.3cm]{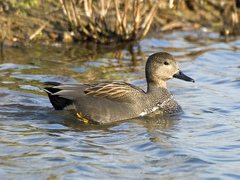} &        
\includegraphics[width=2.3cm]{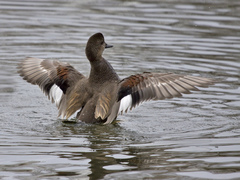} &
\includegraphics[width=2.3cm]{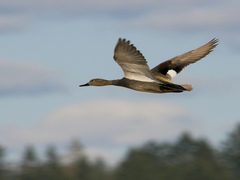} \\
\begin{tabular}[c]{@{}c@{}}Gadwall: 0.65\\ Nighthawk: 0.06\\ Horned Lark: 0.05 \\ $\;$ \end{tabular} &
\begin{tabular}[c]{@{}c@{}}Gadwall: 0.31\\ B. Swallow: 0.14\\ N. Flicker: 0.12 \\ $\;$ \end{tabular} &
\begin{tabular}[c]{@{}c@{}}Gadwall: 0.14\\ Chuck. Widow: 0.13\\ Swain. Warbler: 0.1 \\ $\;$ \end{tabular} & &
\begin{tabular}[c]{@{}c@{}}\underline{Gadwall}: \hlg{0.95}\\ Mallard: 0.01\\ Rb. Merganser: 0.00 \\ $\;$ \end{tabular} &
\begin{tabular}[c]{@{}c@{}}\underline{Gadwall}: \hlg{0.44}\\ Mallard: 0.15\\ Rb. Merganser: 0.11 \\ $\;$ \end{tabular} &
\begin{tabular}[c]{@{}c@{}}Pom. Jaeger: \hlr{0.16}\\ Black Tern: 0.11\\ Herring Gull: 0.07 \\ $\;$ \end{tabular} \\

\includegraphics[width=2.3cm]{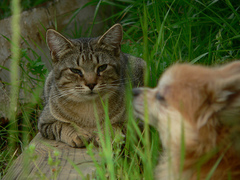} &
\includegraphics[width=2.3cm]{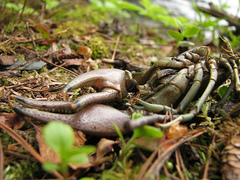} &         
\includegraphics[width=2.3cm]{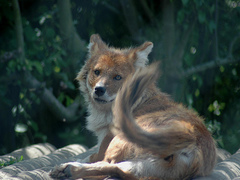} &  &      
\includegraphics[width=2.3cm]{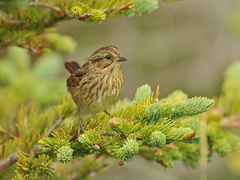} &        
\includegraphics[width=2.3cm]{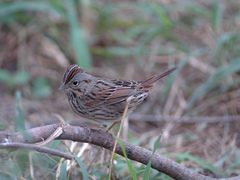} &
\includegraphics[width=2.3cm]{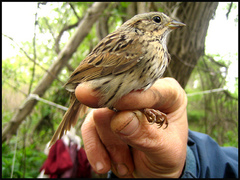} \\
\begin{tabular}[c]{@{}c@{}}Lin. Sparrow: 0.82\\ Ovenbird: 0.07\\ House Sparrow: 0.03 \\ $\;$ \end{tabular} &
\begin{tabular}[c]{@{}c@{}}Lin. Sparrow: 0.49\\ Mockingbird: 0.18\\ N. Waterthrush: 0.07 \\ $\;$ \end{tabular} &
\begin{tabular}[c]{@{}c@{}}Lin. Sparrow: 0.32\\ Song Sparrow: 0.06\\ Tree Sparrow: 0.05 \\ $\;$ \end{tabular} & &
\begin{tabular}[c]{@{}c@{}}\underline{Lin. Sparrow}: \hlg{0.64}\\ Hen. Sparrow: 0.05\\ Clay c. Sparrow: 0.03 \\ $\;$ \end{tabular} &
\begin{tabular}[c]{@{}c@{}}\underline{Lin. Sparrow}: \hlg{0.50}\\ Song Sparrow: 0.24\\ Hen. Sparrow: 0.04 \\ $\;$ \end{tabular} &
\begin{tabular}[c]{@{}c@{}}Hen. Sparrow: \hlr{0.28}\\ Ovenbird: 0.11\\ \underline{Lin. Sparrow}: 0.10 \\ $\;$ \end{tabular} \\

\includegraphics[width=2.3cm]{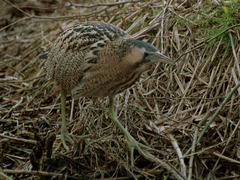} &
\includegraphics[width=2.3cm]{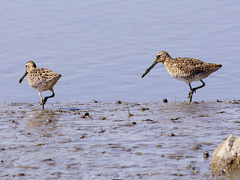} &         
\includegraphics[width=2.3cm]{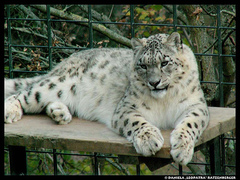} &  &      
\includegraphics[width=2.3cm]{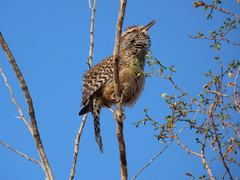} &        
\includegraphics[width=2.3cm]{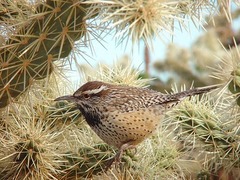} &
\includegraphics[width=2.3cm]{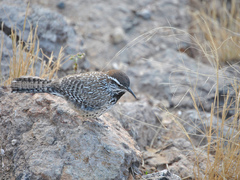} \\
\begin{tabular}[c]{@{}c@{}}Cactus Wren: 0.95\\ W. Meadowlark: 0.02\\ Lin. Sparrow: 0.00 \end{tabular} &
\begin{tabular}[c]{@{}c@{}}Cactus Wren: 0.88\\ Geococcyx: 0.04\\ W. Meadowlark: 0.03 \end{tabular} &
\begin{tabular}[c]{@{}c@{}}Cactus Wren: 0.33\\ N. Flicker: 0.28\\ Lin. Sparrow: 0.07 \end{tabular} & &
\begin{tabular}[c]{@{}c@{}}\underline{Cactus Wren}: \hlg{0.86}\\ Rock Wren: 0.02\\ R. Blackbird: 0.01 \end{tabular} &
\begin{tabular}[c]{@{}c@{}}\underline{Cactus Wren}: \hlg{0.82}\\ N. Flicker: 0.02\\ Geococcyx: 0.02 \end{tabular} &
\begin{tabular}[c]{@{}c@{}}Geococcyx: \hlr{0.25}\\ \underline{Cactus Wren}: 0.20\\ Nighthawk: 0.08 \end{tabular} \\ \bottomrule

\end{tabular}
\caption{\textbf{Qualitative results: \modelname{CUBS200}.} Extends Figure 6 in the main paper.}
\label{tab:expt_qual_cubs}
\end{figure*}

\begin{figure*}[]
\footnotesize
\centering
\begin{tabular}{ccccccc}
\toprule
\multicolumn{3}{c}{{\normalsize Transfer Set}} &  & \multicolumn{3}{c}{{\normalsize Test Set}} \\ \cmidrule{1-3} \cmidrule{5-7}
\includegraphics[width=2.3cm]{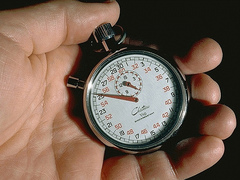} &
\includegraphics[width=2.3cm]{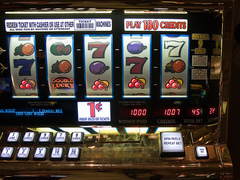} &
\includegraphics[width=2.3cm]{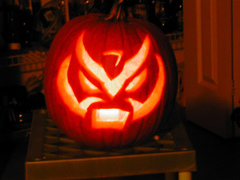} & &
\includegraphics[width=2.3cm]{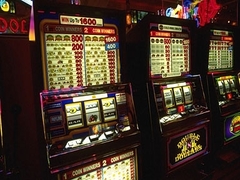} &
\includegraphics[width=2.3cm]{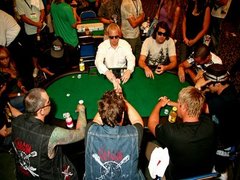} &
\includegraphics[width=2.3cm]{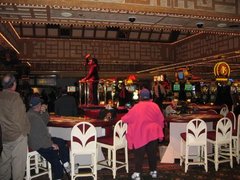} \\
\begin{tabular}[c]{@{}c@{}}Casino: 0.79\\ Deli: 0.03\\ Jewel. Shop: 0.03 \\ $\;$ \end{tabular} &
\begin{tabular}[c]{@{}c@{}}Casino: 0.38\\ Airport ins.: 0.35\\ Bowling: 0.05 \\ $\;$ \end{tabular} &
\begin{tabular}[c]{@{}c@{}}Casino: 0.18\\ Bar: 0.18\\ Movie theater: 0.13 \\ $\;$ \end{tabular} & &
\begin{tabular}[c]{@{}c@{}}\underline{Casino}: \hlg{0.99}\\ Deli: 0.00\\ Toystore: 0.00 \\ $\;$ \end{tabular} &
\begin{tabular}[c]{@{}c@{}}\underline{Casino}: \hlg{0.88}\\ Toystore: 0.08\\ Bar: 0.01 \\ $\;$ \end{tabular} &
\begin{tabular}[c]{@{}c@{}}Restaurant: \hlr{0.46}\\ Bar: 0.24\\ Airport ins.: 0.07 \\ $\;$ \end{tabular} \\

\includegraphics[width=2.3cm]{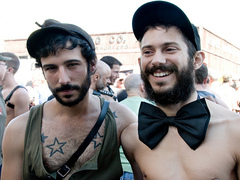} &
\includegraphics[width=2.3cm]{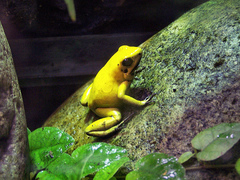} &
\includegraphics[width=2.3cm]{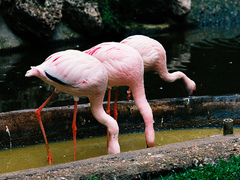} & &
\includegraphics[width=2.3cm]{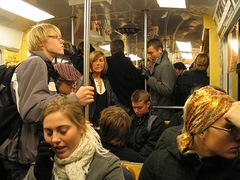} &
\includegraphics[width=2.3cm]{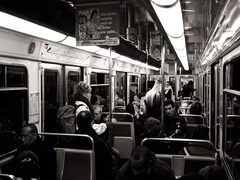} &
\includegraphics[width=2.3cm]{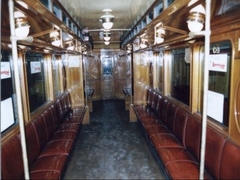} \\
\begin{tabular}[c]{@{}c@{}}Ins. Subway: 0.87\\ Video store: 0.11\\ Dental office: 0.01 \\ $\;$ \end{tabular} &
\begin{tabular}[c]{@{}c@{}}Ins. Subway: 0.59\\ Florist: 0.12\\ Greenhouse: 0.06 \\ $\;$ \end{tabular} &
\begin{tabular}[c]{@{}c@{}}Ins. Subway: 0.37\\ Airport ins.: 0.12\\ Train station: 0.08 \\ $\;$ \end{tabular} & &
\begin{tabular}[c]{@{}c@{}}\underline{Ins. Subway}: \hlg{0.96}\\ Casino: 0.02\\ Museum: 0.01 \\ $\;$ \end{tabular} &
\begin{tabular}[c]{@{}c@{}}\underline{Ins. Subway}: \hlg{0.86}\\ Train station: 0.07\\ Subway: 0.05 \\ $\;$ \end{tabular} &
\begin{tabular}[c]{@{}c@{}}Corridor: \hlr{0.45}\\ \underline{Ins. Subway}: 0.11\\ Bar: 0.09 \\ $\;$ \end{tabular} \\

\includegraphics[width=2.3cm]{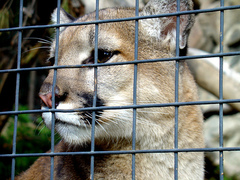} &
\includegraphics[width=2.3cm]{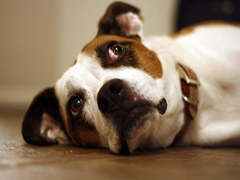} &
\includegraphics[width=2.3cm]{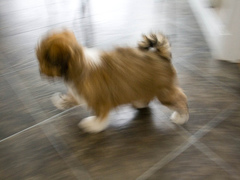} & &
\includegraphics[width=2.3cm]{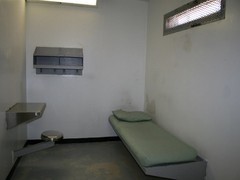} &
\includegraphics[width=2.3cm]{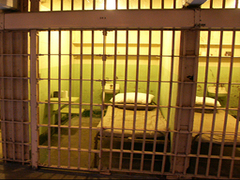} &
\includegraphics[width=2.3cm]{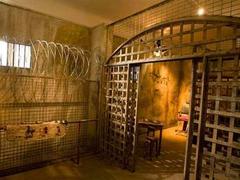} \\
\begin{tabular}[c]{@{}c@{}}Prison cell: 0.52\\ Elevator: 0.20\\ Airport ins.: 0.11 \end{tabular} &
\begin{tabular}[c]{@{}c@{}}Prison cell: 0.23\\ Museum: 0.19\\ Nursery: 0.17 \end{tabular} &
\begin{tabular}[c]{@{}c@{}}Prison cell: 0.21\\ Museum: 0.12\\ Airport ins.: 0.11 \end{tabular} & &
\begin{tabular}[c]{@{}c@{}}\underline{Prison cell}: \hlg{0.83}\\ Kitchen: 0.03\\ Locker room: 0.03 \end{tabular} &
\begin{tabular}[c]{@{}c@{}}\underline{Prison cell}: \hlg{0.52}\\ Subway: 0.08\\ Nursery: 0.07 \end{tabular} &
\begin{tabular}[c]{@{}c@{}}Wine cellar: \hlr{0.31}\\ \underline{Prison cell}: 0.17\\ Staircase: 0.09 \end{tabular} \\ \bottomrule

\end{tabular}
\caption{\textbf{Qualitative results: \modelname{Indoor67}.} Extends Figure 6 in the main paper.  \underline{GT} labels are underlined, \hlg{correct} top-1 knockoff predictions in green and \hlr{incorrect} in red.}
\label{tab:expt_qual_indoor}
\end{figure*}

\begin{figure*}[]
\footnotesize
\centering
\begin{tabular}{ccccccc}
\toprule
\multicolumn{3}{c}{{\normalsize Transfer Set}} &  & \multicolumn{3}{c}{{\normalsize Test Set}} \\ \cmidrule{1-3} \cmidrule{5-7}
\includegraphics[width=2.3cm]{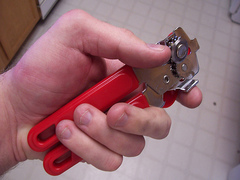} &
\includegraphics[width=2.3cm]{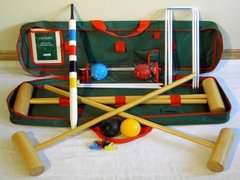} &
\includegraphics[width=2.3cm]{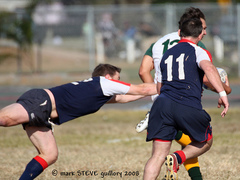} & &
\includegraphics[width=2.3cm]{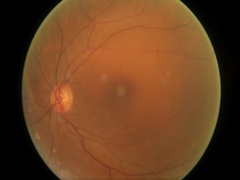} &        
\includegraphics[width=2.3cm]{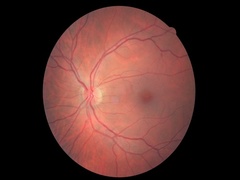} &
\includegraphics[width=2.3cm]{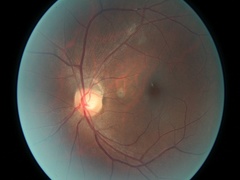} \\
\begin{tabular}[c]{@{}c@{}}No DR: 0.73\\ Proliferative: 0.12\\ Moderate: 0.08 \\ $\;$ \end{tabular} &
\begin{tabular}[c]{@{}c@{}}No DR: 0.48\\ Mild: 0.36\\ Moderate: 0.16 \\ $\;$ \end{tabular} &
\begin{tabular}[c]{@{}c@{}}No DR: 0.30\\ Moderate: 0.29\\ Proliferative: 0.28 \\ $\;$ \end{tabular} & &
\begin{tabular}[c]{@{}c@{}}\underline{No DR}: \hlg{0.50}\\ Mild: 0.33\\ Moderate: 0.13 \\ $\;$ \end{tabular} &
\begin{tabular}[c]{@{}c@{}}\underline{No DR}: \hlg{0.36}\\ Mild: 0.33\\ Moderate: 0.28 \\ $\;$ \end{tabular} &
\begin{tabular}[c]{@{}c@{}}Mild: \hlr{0.53}\\ \underline{No DR}: 0.43\\ Moderate: 0.03 \\ $\;$ \end{tabular} \\

\includegraphics[width=2.3cm]{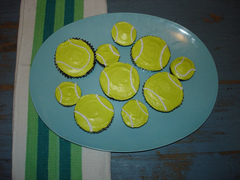} &
\includegraphics[width=2.3cm]{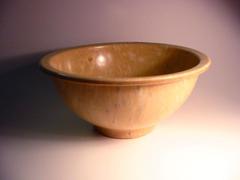} &
\includegraphics[width=2.3cm]{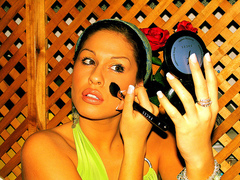} & &
\includegraphics[width=2.3cm]{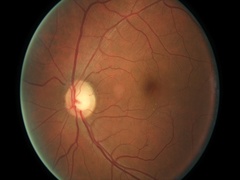} &        
\includegraphics[width=2.3cm]{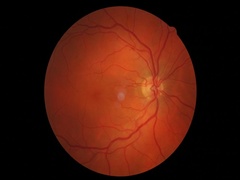} &
\includegraphics[width=2.3cm]{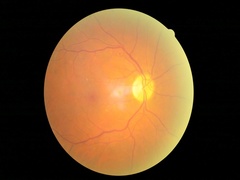} \\
\begin{tabular}[c]{@{}c@{}}Moderate: 0.69\\ No DR: 0.31\\ Mild: 0.01 \\ $\;$ \end{tabular} &
\begin{tabular}[c]{@{}c@{}}Moderate: 0.63\\ No DR: 0.15\\ Severe: 0.13 \\ $\;$ \end{tabular} &
\begin{tabular}[c]{@{}c@{}}Moderate: 0.35\\ Mild: 0.32\\ No DR: 0.22 \\ $\;$ \end{tabular} & &
\begin{tabular}[c]{@{}c@{}}\underline{Moderate}: \hlg{0.48}\\ Mild: 0.316\\ No DR: 0.21 \\ $\;$ \end{tabular} &
\begin{tabular}[c]{@{}c@{}}\underline{Moderate}: \hlg{0.35}\\ Mild: 0.31\\ No DR: 0.23 \\ $\;$ \end{tabular} &
\begin{tabular}[c]{@{}c@{}}No DR: \hlr{0.36}\\ Mild: 0.33\\ \underline{Moderate}: 0.26 \\ $\;$ \end{tabular} \\

\includegraphics[width=2.3cm]{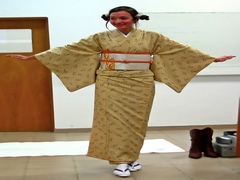} &
\includegraphics[width=2.3cm]{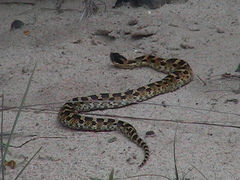} &
\includegraphics[width=2.3cm]{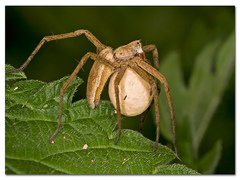} & &
\includegraphics[width=2.3cm]{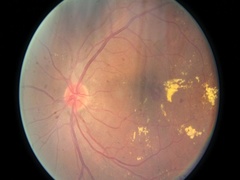} &        
\includegraphics[width=2.3cm]{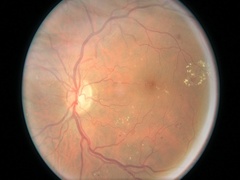} &
\includegraphics[width=2.3cm]{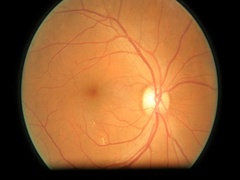} \\
\begin{tabular}[c]{@{}c@{}}Severe: 0.73\\ Proliferative: 0.23\\ Moderate: 0.04 \end{tabular} &
\begin{tabular}[c]{@{}c@{}}Severe: 0.70\\ Proliferative: 0.30\\ Moderate: 0.00 \end{tabular} &
\begin{tabular}[c]{@{}c@{}}Severe: 0.53\\ Mild: 0.16\\ Moderate: 0.15 \end{tabular} & &
\begin{tabular}[c]{@{}c@{}}\underline{Severe}: \hlg{0.57}\\ Moderate: 0.23\\ Proliferative: 0.19 \end{tabular} &
\begin{tabular}[c]{@{}c@{}}\underline{Severe}: \hlg{0.41}\\ Proliferative: 0.29\\ Moderate: 0.24 \end{tabular} &
\begin{tabular}[c]{@{}c@{}}Moderate: \hlr{0.62}\\ \underline{Severe}: 0.16\\ Mild: 0.13 \end{tabular} \\ \bottomrule
\end{tabular}
\caption{\textbf{Qualitative results: \modelname{Diabetic5}.} Extends Figure 6 in the main paper.}
\label{tab:expt_qual_diabetic}
\end{figure*}